\documentclass[10pt,twocolumn,letterpaper]{article}

\usepackage[pagenumbers]{iccv} 

\usepackage[accsupp]{axessibility}

\usepackage{xspace}
\usepackage{graphicx}
\usepackage{amsmath}
\usepackage{amssymb}
\usepackage{amsthm}
\usepackage{adjustbox}
\usepackage{algorithm}
\usepackage{algpseudocode}
\usepackage{wrapfig}
\usepackage{lipsum}
\usepackage{booktabs}
\usepackage{multirow}
\usepackage{makecell}

\newcommand{\name}[0]{FrameFusion\xspace}
\newcommand{\xhdr}[1]{{\noindent\bfseries #1}.}

\newcommand{\rom}[1]{\textup{\uppercase\expandafter{\romannumeral#1}}}

\newcounter{observation}
\newenvironment{observation}[1][]{\refstepcounter{observation}\par\medskip
   \noindent \textbf{Observation~\theobservation. #1} \rmfamily}{\medskip}

\newcounter{design}
\newenvironment{design}[1][]{\refstepcounter{design}\par\medskip
   \noindent \textbf{Design Choice~\thedesign. #1} \rmfamily}{\medskip}

\usepackage[most]{tcolorbox}
\usepackage{colortbl}
\usepackage{pifont}
\usepackage{makecell}

\newtcolorbox[
  auto counter,
  list inside=examplelist,
]{greenbox}[2][]{
  enhanced,
  title={Example~\thetcbcounter. \textbf{#1}},
  colback=green!5!white,
  colframe=green!50!black,
  colbacktitle=green!60!black,
  coltitle=white,
  #2
}

\newtcolorbox[
  auto counter,
  list inside=examplelist
]{bluebox}[2][]{
  enhanced,
  title={Example~\thetcbcounter. \textbf{#1}},
  colback=cyan!5!white,  
  colframe=cyan!50!black,  
  colbacktitle=cyan!75!black,  
  coltitle=white,
  #2
}

\usepackage{pifont}
\definecolor{iccvblue}{rgb}{0.21,0.49,0.74}
\usepackage[pagebackref,breaklinks,colorlinks,allcolors=iccvblue]{hyperref}

\def\paperID{7346} 
\def\confName{ICCV}
\def\confYear{2025}

\title{FrameFusion: Combining Similarity and Importance for Video Token Reduction on Large Vision Language Models}

\author{%
Tianyu Fu\thanks{Equal contribution.}$^{~ 1,2}$,
Tengxuan Liu$^{* 1,2}$,
Qinghao Han$^{* 3}$,
\\
Guohao Dai$^{4,2}$, 
Shengen Yan$^{2}$, 
Huazhong Yang$^{1}$, 
Xuefei Ning$^{1}$,
Yu Wang$^{1}$
\\
{\normalsize
$^{1}$Tsinghua University
$^{2}$Infinigence-AI
$^{3}$Peking University
$^{4}$Shanghai Jiao Tong University
}
}

\begin{document}
\maketitle
\begin{abstract}
The increasing demand to process long and high-resolution videos significantly burdens Large Vision-Language Models (LVLMs) due to the enormous number of visual tokens.
Existing token reduction methods primarily prune tokens based on importance metrics, such as cumulative attention scores. However, even important tokens may exhibit high redundancy caused by similarity among adjacent video frames and repetitive visual elements.
To address this limitation, we propose \name, a novel token reduction approach integrating similarity-based merging with importance-based pruning.
We conduct a thorough study on token similarity characteristics, revealing three key insights: (1) spatially corresponding visual tokens between adjacent frames have higher cosine similarities compared to other token pairs; (2) high token similarities prominently decrease in deeper model layers; and (3) token similarity rankings are highly consistent across different layers.
Guided by these observations, \name computes token similarities exclusively between corresponding visual tokens from adjacent frames, applies token merging at initial successive layers followed by pruning in deeper layers, and adopts a cascaded merging strategy to further enhance efficiency.
We evaluate \name comprehensively across six diverse LVLMs, ranging from 2B to 72B parameters, using five video benchmarks encompassing video retrieval, question-answering, and spatial-temporal understanding tasks. 
Experiments show that \name reduces visual tokens by 70\%, achieving 1.6 – 3.6$\times$ end-to-end speedups, with an average performance impact of less than 3\%.
Our code is available at \url{https://github.com/thu-nics/FrameFusion}.
\end{abstract}    
\section{Introduction}
\label{sec:intro}
Large Vision-Language Models (LVLMs) have demonstrated remarkable capabilities across various video understanding tasks, including temporal and spatial perception, recognition, and reasoning~\citep{llava-vid, minicpm, li2024llava-onevision, chen2024internvl}.
Increasingly demanding applications require LVLMs to process longer and more complex videos~\citep{team2024gemini, chai2024auroracap, li2025llama-vid, weng2025longvlm}.

However, handling extensive video data incurs substantial computational overhead. 
Typically, LVLMs sample frames from videos, divide each frame into image patches, and embed these sequentially as visual tokens through a visual encoder. While effective, this process generates an enormous number of tokens. For instance, Google's Gemini, with a standard sampling rate of 1 frame per second (fps), needs to process approximately one million tokens to analyze an hour-long video~\citep{team2024gemini}.

\begin{figure}[tb]
    \centering
    \includegraphics[width=\linewidth]{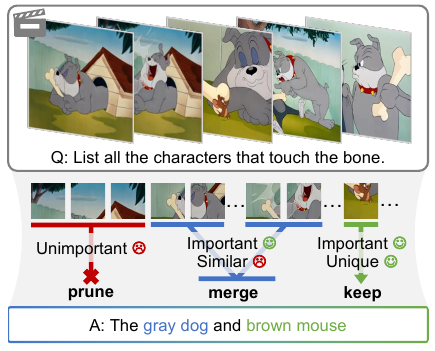}
    \vspace{-16pt}
    \caption{The central idea of \name. Compared with importance-based token pruning, \name additionally applies similarity-based token merging, keeping only important and unique visual tokens.}
    \vspace{-12pt}
    \label{fig:illustration}
\end{figure}

Previous works primarily employ importance-based token pruning methods to mitigate efficiency demands. 
These approaches reduce visual tokens based on metrics such as cumulative attention scores~\citep{Wang2020SpAtten, Zhang2023H2O, fu2024lazyllm} or normalized token feature~\citep{chen2023sparsevit}. 
Nevertheless, among the top 10\% of tokens ranked by cumulative attention scores, 55\% of them exhibit high redundancy, with a cosine similarity above 0.9 in the first layer of Llava-Video-7B.

In this work, we revisit token similarity as a perpendicular factor to token importance for reducing visual tokens.
We argue that even important tokens can introduce redundancy due to visual similarities, particularly among adjacent frames. By merging these highly similar tokens, redundancy can be significantly reduced without compromising essential visual information.

To effectively utilize token similarity, we first systematically investigate its characteristics within LVLMs. We find that (1) token similarity predominantly occurs between spatially corresponding tokens from adjacent frames, (2) similarities exhibit high values particularly at shallow layers, and (3) the token similarity rankings are highly consistent across layers.

Based on these insight, we propose \name, a plug-and-play approach that integrates similarity-based merging with importance-based pruning. \name efficiently computes token similarities, progressively merges similar tokens at shallow layers, and subsequently applies importance-based pruning to adhere to computational constraints.
Our contributions are summarized as follows:
\begin{enumerate}
    \item We systematically analyze token similarity characteristics across input positions and layers in LVLMs.
    \item We propose \name, a novel, post-training method that integrates similarity-based token merging with importance-based pruning for video token reduction.
    \item We validate the effectiveness of \name through extensive experiments across diverse LVLMs, model sizes, input lengths, and video benchmarks.
\end{enumerate}

Experiments confirm that \name effectively advances the Pareto front in token compression. 
It reduces visual tokens to 30\%, achieving 1.6–3.6$\times$ end-to-end speedup while maintaining less than a 3\% performance drop over dense models. Its simple and effective design ensures broad applicability across various tasks and scenarios.
\section{Related Work}
\label{sec:related_work}

\subsection{Large Vision Language Model (LVLMs)}
The LVLM architecture typically consists of a visual encoder and a Large Language Model (LLM)~\citep{llava-vid, minicpm, li2024llava-onevision, chen2024internvl, li2025llama-vid, weng2025longvlm, efficient_survey}. 
The visual encoder converts visual inputs into token sequences, which the LLM then processes alongside text sequences to generate responses.
Specifically, for video input, frames are first sampled temporally and then specially divided into sequences of image patches before sending to the visual encoder~\citep{llava-vid, li2024llava-onevision, minicpm}, as shown in Figure~\ref{fig:illustration}.
Due to the high temporal and spatial resolution demands of complex video understanding tasks, token lengths can reach up to one million for an hour-long video~\citep{team2024gemini}, imposing significant computational overhead on LVLMs.

\subsection{Token Compression}
Motivated by the heavy overhead of video processing, token compression becomes an essential method for LVLM efficiency. 
Existing methods compress tokens at three subsequent processes.

The first branch of work reduces the initial input before sending them to visual encoder.
They set rules to mix different temporal sampling frequencies~\citep{xu2024slowfast, llava-vid} and special resolutions~\citep{minicpm, dong2024internlm-xcomposer2} when converting videos to input sequences, introducing trade-offs between visual detail and efficiency.
Despite simplicity, they inevitably incur direct detail losses and neglect the content guidance for compression.

Other works reduce tokens inside the visual encoder. 
They selectively retrieve~\citep{he2024ma-llm} or condense~\citep{li2025llama-vid} visual tokens in the visual encoder.
Yet, they require re-encoding all visual tokens if the text instruction changes, which incurs significant overheads for common multi-round conversation scenarios. Besides, an additional model fine-tuning is often needed to align the new visual encoding space~\citep{Jian2023EVLGen}.

Another branch of work focuses on token reduction in the subsequent LLM.
For text-only tasks, previous works design static~\citep{fu2024moa, xiao2023streamingLLM, han2023lmInfinite} or dynamic~\citep{jiang2024minference, Zhang2023H2O, Ge2023FastGen, Liu2023Scissorhands} pruning pattern based on the importance of token (or KV-Cache).
Emerging concurrent works highlight the specific token importance distribution for vision-language tasks~\citep{chen2024fastv, li2025flexattention, tu2024vl-cache, zhang2024sparsevlm, zhang2024avl}, which further increase the sparsity of importance-based token pruning.
However, as shown in Figure~\ref{fig:spearman_rank}, token importance is inconsistent across different layers. It incurs prediction loss by pruning an unimportant token at shallow layers, which becomes important but inaccessible at deeper layers.
\name falls in this category, exploring a more consistent and perpendicular token reduction method: similarity-based token merging. More related works are discussed in Appendix \ref{sec:appendix/related_work}.

\section{Token Similarity Analysis}
\label{sec:analysis}
While token importance in LVLMs has been extensively explored~\citep{chen2024fastv, li2025flexattention, tu2024vl-cache, zhang2024sparsevlm, zhang2025spargeattn}, the characteristics of token similarity remain under-investigated. To bridge this gap, we conduct comprehensive oracle experiments analyzing token similarity and contrasting it with token importance.

\subsection{Experimental Setup and Definitions}
\label{sec:analysis/notion}
In this section, we present oracle experiment results on the Llava-Video-7B~\citep{llava-vid} model using the first 128 video samples from the VideoMME dataset~\citep{fu2024videomme}, each comprising 64 frames sampled at 1 fps. Metrics reported are averaged over all samples unless otherwise noted. Similar results on other models are included in Appendix~\ref{sec:appendix/observation}.

We define token importance and token similarity as $I^{(l)} \in \mathbb{R}^N$ and $S^{(l)} \in \mathbb{R}^N$, respectively, with $l$ indicating the LLM layer index and $N$ denoting the number of input tokens. 
We use subscript $t$ to index the token along the input length dimension $N$.
For simplicity, we omit the layer index when contextually clear. 
Input hidden features for layer $l$ are denoted as $\mathbf{X}^{(l)} \in \mathbb{R}^{N \times d}$.

Following previous works~\citep{chen2024fastv, Zhang2023H2O}, token importance $I^{(l)}_t$ is computed using the cumulative attention score, calculated by summing the post-softmax attention scores vertically across the $t$-th column and averaging this sum across all attention heads at layer $l$. 

For token similarity $S_t^{(l)}$, we define it specifically between each visual token and its spatially corresponding token from the preceding frame, based on our empirical observation detailed in Section~\ref{sec:analysis/where_when_similar}. 
Formally, token similarity is computed with cosine similarity:
\begin{equation}
S_t = \frac{X_{t-P}^T X_t}{\lVert X_{t-P} \rVert_2 \cdot \lVert X_t \rVert_2},
\label{eq:token_similarity}
\end{equation}
where $P$ represents the number of visual tokens per frame, and $T$ represents the matrix transpose.


\subsection{Where Does High Similarity Occur?}
\label{sec:analysis/where_when_similar}

\begin{figure}
    \centering
    \includegraphics[width=0.68\linewidth]{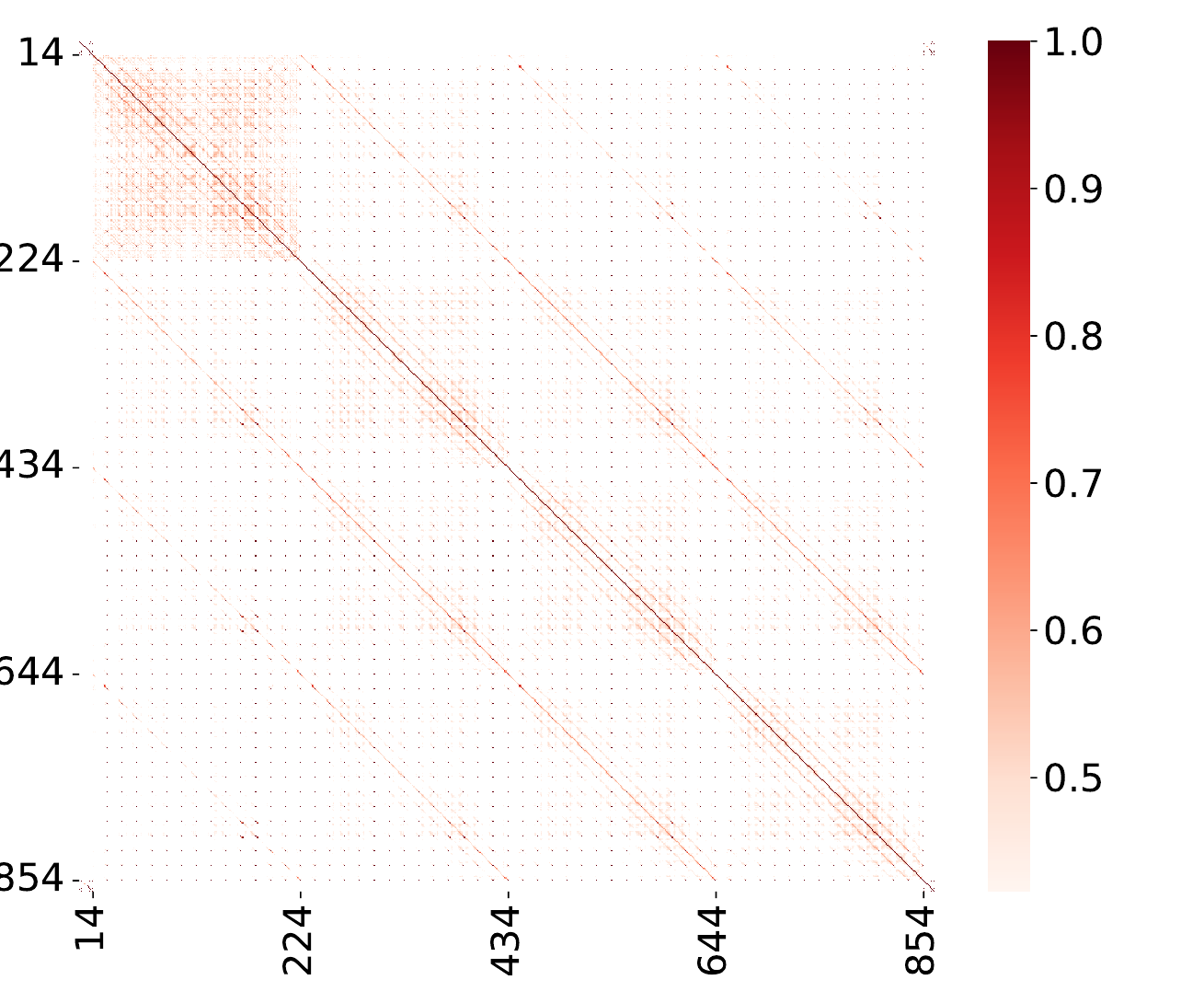}
    \vspace{-8pt}
    \caption{Token similarities among all input tokens at the first LVLM layer in Llava-Video-7B models. For visual clarity, the color bar displays only the top 90\% of similarity values. Visual tokens begin at index 14, with 210 tokens per frame.}
    \label{fig:token_similarity}
\end{figure}

We first analyze which tokens typically exhibit high similarity, as these tokens have greater potential for merging.
To maintain visual clarity, we limit the video frames to 4 (resulting in 840 visual tokens at $P=210$ tokens per frame), with 14 system prompt tokens and 20 user instruction tokens before and after the visual tokens.
Figure~\ref{fig:token_similarity} illustrates the $N\times N$ cosine similarity matrix at the first LVLM layer.
A distinct 210th sub-diagonal emerges, highlighting significantly higher similarities between tokens at positions $i$ and $i+P$. Statistically, the average cosine similarity of 0.62 between these tokens far exceeds the average similarity (0.28) observed elsewhere. Hence, we conclude:
\begin{observation}
Spatially corresponding visual tokens from adjacent frames exhibit higher cosine similarity compared to other token pairs.
\label{obs:adjacent}
\end{observation}

Based on this observation, we focus on similarities among these particular tokens and define token similarity as described in Equation~\ref{eq:token_similarity}. Additionally, sub-diagonals at multiples of 210 tokens also show relatively high similarities. This indicates that visual redundancy extends across multiple consecutive frames, which can also be identified by sequentially examining adjacent frames.


\subsection{What Is the Token Similarity Distribution Across Layers?}
\label{sec:analysis/distribution}

\begin{figure}
\centering
\includegraphics[width=\linewidth]{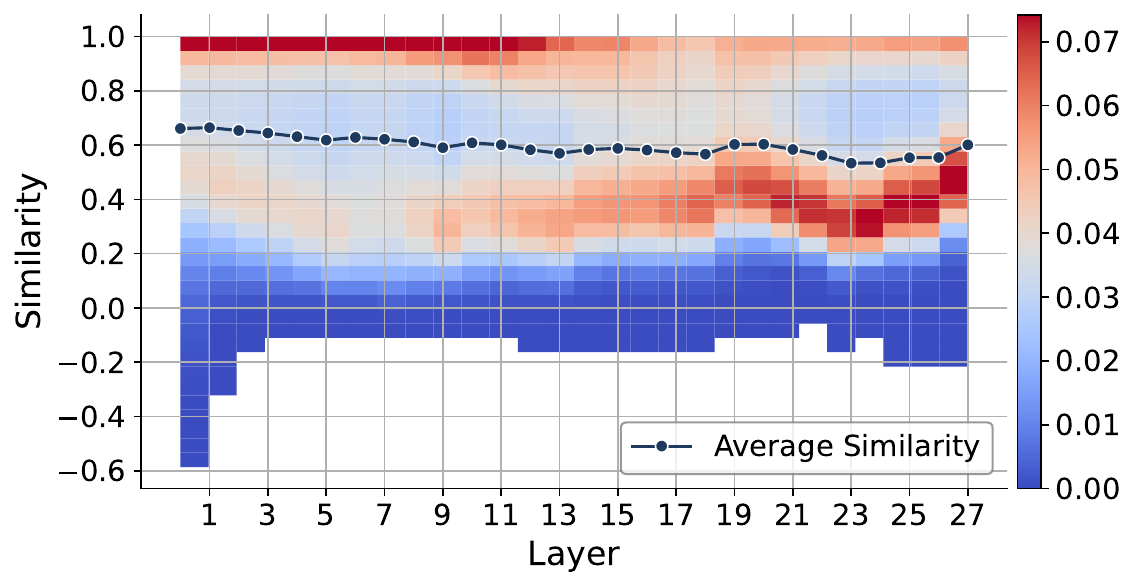}
\vspace{-8pt}
\caption{
Heatmap of token similarity across model layers. 
Each cell represents a similarity range at a specific layer, with color intensity denoting distribution frequency.
The line overlay shows the mean token similarity per layer.
}
\label{fig:similarity_distribution}
\end{figure}

To identify the most effective layers for token compression, we analyze the distribution of token similarity values across model layers. 
Figure~\ref{fig:similarity_distribution} shows that, although the mean token similarity remains relatively stable across layers, the distribution shifts noticeably:
\begin{observation}
High token similarity values decrease in deeper model layers.
\label{obs:shift}
\end{observation}

This trend is particularly evident in Llava-Video model. While the mean similarity decreases only marginally (from 0.62 to 0.60), the distribution significantly condenses as layers deepen. Specifically, the 30th percentile similarity value decreases from 0.90 at the first layer to 0.72 at the last.
Additional numerical results are in Appendix~\ref{sec:appendix/similarity_distribution}.

The causal attention mechanism in LLMs contributes to this shift because tokens at later positions can aggregate information from earlier tokens, but not vice versa.
This directional aggregation causes initially similar tokens, such as those corresponding across adjacent frames, to diverge increasingly at deeper layers.

Given this observation, \name prioritizes token merging at shallower layers to effectively leverage these initially high similarities.

\subsection{Is Token Similarity Ranking Consistent Across Layers?}
The consistency of token similarity rankings across layers determines whether tokens that are similar at shallow layers remain similar in deeper layers.

\begin{figure}
    \centering
    \includegraphics[width=\linewidth]{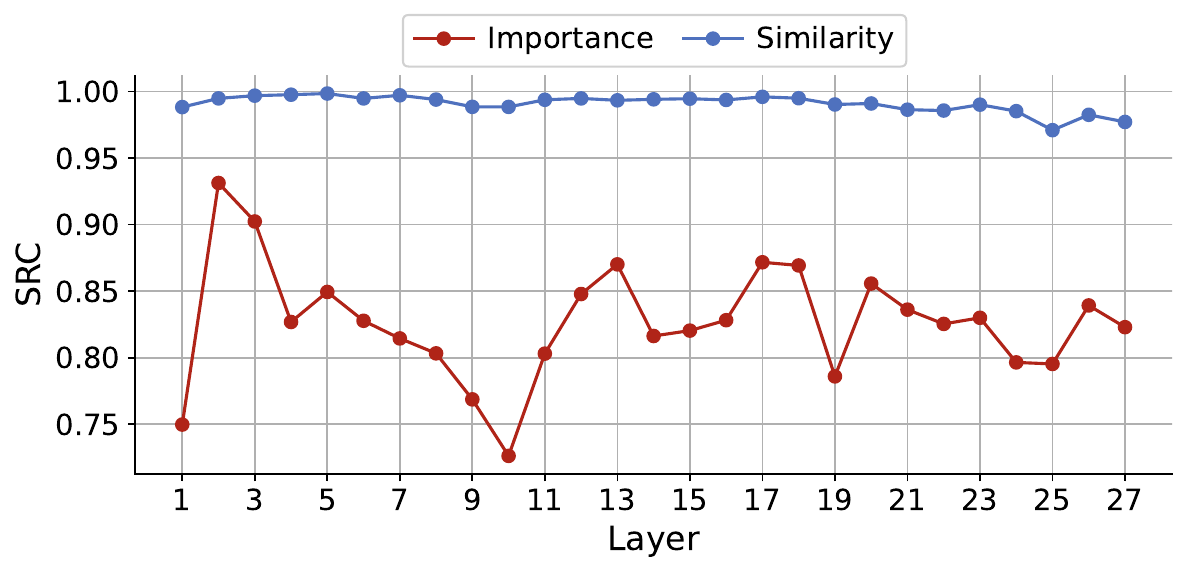}
    \vspace{-12pt}
    \caption{Spearman Rank Correlation (SRC) between adjacent layers for the Llava-Video-7B model.}
    \label{fig:spearman_rank}
\end{figure}

We first quantify the consistency of token similarity rankings between adjacent layers using \textit{Spearman Rank Correlation (SRC)}~\citep{zar2005spearman}, which measures the correlation of rankings across different layers. For comparison, we also compute the SRC for token importance. 
As shown in Figure\ref{fig:spearman_rank}, token similarity maintains consistently high SRC values approaching 1, indicating stable rankings across layers. In contrast, token importance shows lower and unstable SRC values, suggesting that tokens considered unimportant at shallow layers might become important at deeper layers.

\begin{figure}[tb]
    \centering
    \begin{subfigure}[b]{\linewidth}
        \centering
        \includegraphics[width=\linewidth]{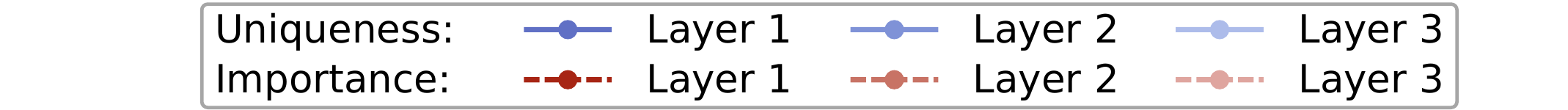}
        \vspace{-13pt}
    \end{subfigure}

    \begin{subfigure}[b]{\linewidth}
        \centering
        \includegraphics[width=\linewidth]{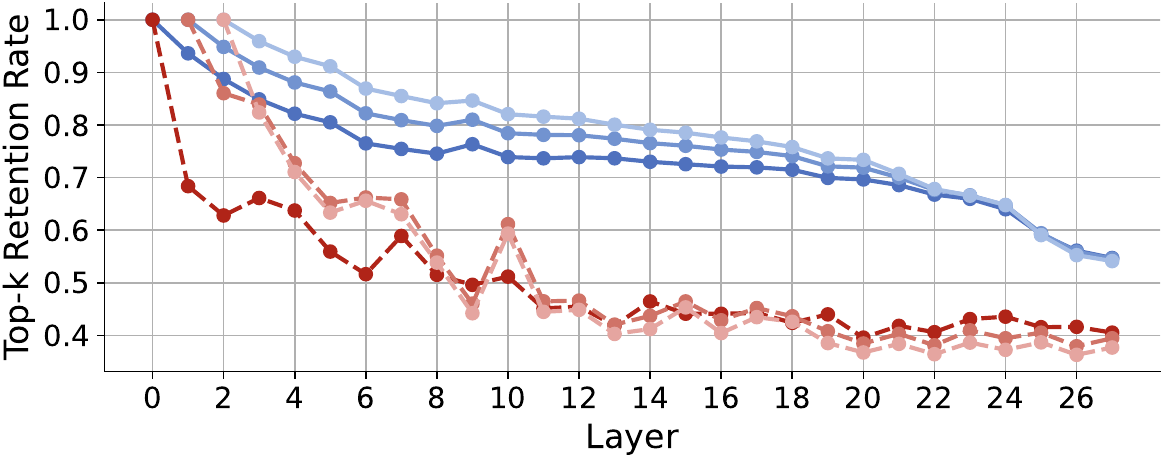}
    \end{subfigure}
    \vspace{-12pt}
    \caption{The top-30\% retention rate across model layers using different retention metrics and starting layers.}
    \label{fig:retention_rate_plot}
\end{figure}

We further examine ranking consistency between shallow and deep layers using the \textit{top-k retention rate}. 
For notion simplicity, we define token uniqueness as 1 minus similarity.
To calculate the retention rate, we identify the top-$k$ tokens at each layer based on token uniqueness or importance. The top-k retention rate at layer $l$ with respect to a starting layer $i$ is defined as the intersection ratio between the top-k tokens at layers $l$ and $i$. Formally:

\begin{equation}
    T^{(l)} = \left\{i \; \big|\; i \in \operatorname{Top-K}(f(\mathbf{X}^{(l)}))\right\},
\end{equation}
where $f$ represents token uniqueness or token similarity, and $\operatorname{Top-K}$ selects indices of the top-k values. The retention rate $R^{(l)}i$ is then calculated as:
\begin{equation}
R^{(l)}{i} = |T^{(l)} \cap T^{(i)}|/|T^{(i)}|.
\end{equation}

As shown in Figure~\ref{fig:retention_rate_plot}, we calculate the top-30\% retention rate at shallow starting layers $0$ to $2$ under different metrics. The token similarity exhibits a much higher retention rate than token importance. Based on these analyses, we conclude:
\begin{observation}
Token similarity rankings are more consistent across layers than token importance rankings.
\label{obs:consistency}
\end{observation}

Combining Observations~\ref{obs:shift} and~\ref{obs:consistency}, we highlight a key phenomenon: while the similarity between highly similar tokens decreases at deeper layers, the relative similarity rankings of these tokens remain stable.
Motivated by this stable ranking, \name merges highly similar tokens at shallow layers, maintaining this reduction throughout subsequent layers.
\section{\name Design}
\label{sec:method}

\begin{figure}
    \centering
    \includegraphics[width=\linewidth]{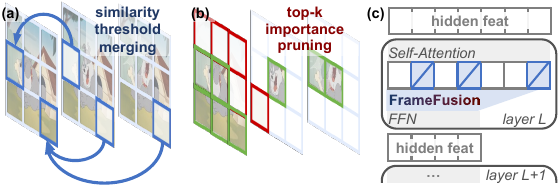}
    \vspace{-8pt}
    \caption{\name first (a) merges tokens with similarities above a specified threshold at shallow layers, then (b) applies top-$k$ importance pruning to comply with the given computational constraints. (c) Tokens are permanently reduced for subsequent layers.}
    \label{fig:method_detail}
\end{figure}

Building upon the observations from Section~\ref{sec:analysis}, we propose \name, a novel token compression method for video LVLMs, exploring the new perspective of token similarity. 
The detailed design is introduced in Section~\ref{sec:method/design}, followed by the rationales behind key design choices in Section~\ref{sec:method/reason}.

\subsection{Two-Stage Token Compression}
\label{sec:method/design}

The core idea of \name is illustrated in Figure~\ref{fig:illustration}. Unlike traditional methods that solely rely on importance-based pruning, \name additionally merges similar tokens before pruning, retaining only those that are both important and unique. The two-stage token compression approach is depicted in Figure~\ref{fig:method_detail}.

\xhdr{Merging stage}
In the first merging stage, \name utilizes token similarity to merge visual tokens. 
Specifically, it computes token similarity $S^{(l)}$ at each shallow layer according to Equation~\ref{eq:token_similarity}, considering only $N$ cosine similarities between corresponding tokens from adjacent frames. Tokens whose similarity exceeds a predefined threshold ($S_{\text{threshold}}$) are grouped with their corresponding tokens from previous frames. These merging groups are transitive, allowing concatenated groups to form a larger group containing more than two tokens (as shown in Figure~\ref{fig:method_detail}). 
Within each group, \name performs element-wise averaging of all tokens and assigns the averaged result to the earliest token in the group. 
This forward merging strategy ensures that subsequent visual tokens can still aggregate information from all preceding tokens using causal attention.

This merging procedure is progressively applied at successive shallow LLM layers until the number of similar tokens falls below a predefined threshold ($N_{\text{threshold}}$). 
Merging is applied before the feed-forward network (FFN) module. However, to further reduce token numbers in the attention module of the first LLM layer, an additional merging step is performed before it. After completing the merging stage, the remaining tokens proceed to the pruning stage.

\xhdr{Pruning stage}
After the merging stage, \name further prunes unimportant tokens. As defined in Section~\ref{sec:analysis/notion}, \name uses cumulative attention scores to represent token importance. 
Given a user-specified computational budget, \name determines the maximum allowable number of tokens ($k$). It then applies pruning to retain only the top-$k$ tokens based on importance scores.

Through the merging and pruning stages, \name effectively retains only unique and important visual tokens for subsequent processing, significantly enhancing the efficiency of LVLMs.
We also investigate the effect of different $S_{\text{threshold}}$ and $N_{\text{threshold}}$.
Experimental results indicate that adjustments to these thresholds result in only minor variations in model performance.
Detailed results are presented in Section~\ref{sec:ablation_threshold}.

\subsection{Design Choice Rationales}
\label{sec:method/reason}

In this subsection, we explain the rationales behind the key design choices of \name, grounded in the observations from Section~\ref{sec:analysis}. More empirical evidences are presented in Section~\ref{sec:ablation}.

\begin{design}
\name computes token similarities only between corresponding visual tokens of adjacent frames.
\label{dsn:corr_token_only}
\end{design}

Unlike token importance, which reuses the existing $N \times N$ attention scores, token similarity introduces a new, orthogonal metric. 
To avoid additional $N \times N$ similarity computations for all token pairs, we exploit Observation~\ref{obs:adjacent} to compute only empirically similar token pairs with an $O(N)$ complexity.

\begin{design}
\name applies token merging at the initial successive layers, followed by pruning at deeper layers.
\label{dsn:merge_then_prune}
\end{design}

Another critical design choice involves determining the appropriate layers for different token reduction methods. 
For importance-based pruning, previous studies indicate a decline in visual token importance after the initial layers~\citep{chen2024fastv}, recommending less pruning at shallow layers~\citep{fu2024moa, cai2024pyramidkv}.
In contrast, similarity-based merging depends on initially high token similarities, preferring shallow layers as per Observation~\ref{obs:shift}. 
Given these contrasting preferences, \name employs merging at shallow layers and pruning at deeper layers to optimize both similarity and importance metrics.

\begin{design}
\name merges tokens in a cascaded manner.
\label{dsn:cascaded_merging}
\end{design}

The final design choice addresses whether merged tokens should remain combined across subsequent layers (cascaded merging) or be individually reconsidered at each layer (non-cascaded merging).

Specifically, as shown in Figure~\ref{fig:method_detail}(c), cascaded merging permanently reduces the token count once tokens are merged, significantly lowering computational costs in both feed-forward network (FFN) and attention modules at subsequent layers.
In contrast, non-cascaded merging maintains the original token count at every layer. It selectively reduces computations within certain modules, typically pruning only the Key and Value matrices in attention layers~\citep{tang2024quest, li2024snapkv, Zhang2023H2O}.
Although non-cascaded merging retains flexibility by potentially reusing tokens at deeper layers, it incurs additional computational overhead due to repeated similarity evaluations and unchanged FFN computations.

Given these accuracy-efficiency trade-offs, the optimal merging strategy depends on whether tokens merged at shallow layers remain similar in deeper layers. Considering the higher consistency of token similarity rankings across layers (Observation~\ref{obs:consistency}), \name employs cascaded merging to eliminate unnecessary computations.

These rationales illustrate the motivation behind each design choice, clarifying the superior performance of \name.
\section{Experiment}

\subsection{Setups}
\xhdr{Baselines}
We compare \name with state-of-the-art token pruning baselines, StreamingLLM~\citep{xiao2023streamingLLM}, FastV~\citep{chen2024fastv}, and PruMerge~\citep{shang2024prumerge}.
Hyperparameters adhere to respective official implementations and are detailed in Appendix~\ref{sec:appendix/experiment_setup}.

\xhdr{Models}
We evaluate our approach across six video LVLMs from diverse model families and sizes, including lmms-lab models: LLaVA-Video-\{7B,72B\}-Qwen2 (denoted Llava-Video-\{7B,72B\})~\citep{llava-vid}; NVLabs models: NVILA-Lite-2B, NVILA-8B-Video, NVILA-Lite-15B-Video (denoted NVILA-\{2B,8B,15B\})~\citep{liu2024nvila}; and OpenBMB model MiniCPM-V-2\_6 (denoted MiniCPM-V-8B)~\citep{minicpm}.
The PruMerge baseline is incompatible with MiniCPM-V due to its Q-Former architecture and is excluded for this model.

\xhdr{Benchmarks}
We use lmms-eval~\citep{zhang2024lmmseval} as the primary evaluation framework and test five video benchmarks: Video Needle In A Haystack (VideoNIAH)\citep{zhao2024VNIAH} for visual content retrieval; NExT-QA~\citep{xiao2021nextqa} for video question-answering; and VideoMME~\citep{fu2024videomme}, EgoSchema~\citep{mangalam2023egoschema} and MVBench~\citep{li2024mvbench} for general video understanding, highlighting spacial and temporal understanding, respectively.

\xhdr{Token Budget}
We define \textit{token budget} as the average sequence length of KV-Cache at the start of the decoding stage.
For cascaded methods (\name, FastV, and PruMerge), it also equals the average token length per layer of the prefill stage. For StreamingLLM, it equals the sink size plus window size.
The \textit{relative token budget}, denoted $C$, is the token budget divided by the original input length $N$. 
Unless specified otherwise, we set $C=30\%$ for all token compression methods.

\subsection{Computation-Accuracy Trade-off}

\begin{figure}[tb]
    \centering
    \begin{subfigure}[b]{\linewidth}
        \centering
        \includegraphics[width=\linewidth]{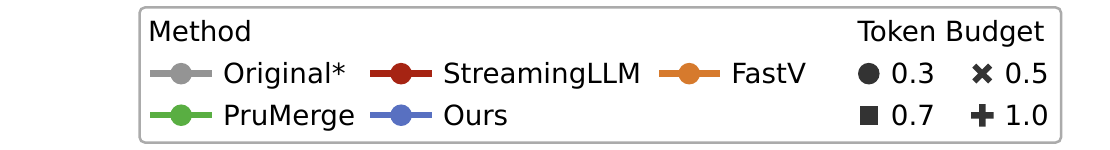}
        \vspace{-14pt}
    \end{subfigure}

    \begin{subfigure}[b]{\linewidth}
        \centering
        \includegraphics[width=\linewidth]{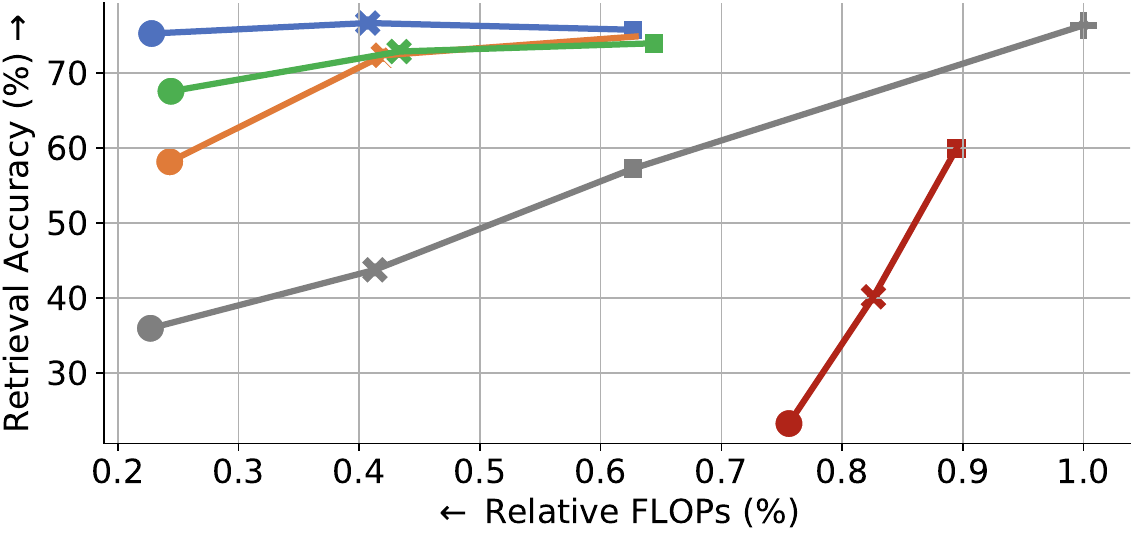}
        \vspace{-15pt}
    \end{subfigure}

    \caption{The accuracy-computation trade-offs of various token compression methods, tested on Llava-Video-7B with VideoNIAH benchmark. Original* represents the original model with reduced frame rates.}
    \label{fig:pareto}
    \vspace{-12pt}
\end{figure}

Figure~\ref{fig:pareto} explores the computation-accuracy trade-offs of \name by varing the relative token budget. 
The x-axis shows the relative computation FLOPs, normalized to the original dense model operating at a 1 frame-per-second (fps) sampling rate. The y-axis shows the VideoNIAH retrieval accuracy.
Higher accuracy at lower FLOPs (towards the top-left) indicates better trade-offs.
The \textit{Original$^*$} baseline, which directly adjusts the sampling rate of the original model, shows that \name achieves faster accuracy gains per FLOP compared to directly increasing frame rates. 
Other baselines also show significant accuracy degradation at reduced token budgets. 
In contrast, \name maintains high accuracy even at 30\% computing FLOPs, greatly advancing the Pareto Front. 
We also explore the computation-accuracy trade-offs on different models and benchmarks.
The results are detailed in Appendix \ref{sec:appendix/trade-off}.

\subsection{Performance}
\label{sec:exp/performance}

\begin{table*}[t]
\centering
\setlength\tabcolsep{3pt}
\begin{tabular}{lll|ccc|cc|cc|cc|c}
\toprule
\multirow{2}{*}{\textbf{Model}} 
 & \multirow{2}{*}{\textbf{Size}} 
 & \multirow{2}{*}{\textbf{Method}} 
 & \multicolumn{3}{c|}{\textbf{VideoNIAH}}
 & \multicolumn{2}{c|}{\textbf{NExT-QA}} 
 & \multicolumn{2}{c|}{\textbf{VideoMME}} 
 & \multirow{2}{*}{\textbf{EgoSchema}}
 & \multirow{2}{*}{\textbf{MVBench}}
 & \multirow{2}{*}{\textbf{Avg.$\uparrow$}}\\
 & & & edit & insert1 & insert2 & mc & oe & w/o sub. & w/ sub. & & & \\
\midrule
\multirow{11}{*}{Llava-Video}& \multirow{5}{*}{7B}
 & Original & 90.7 & 50.7 & 88.0 & 83.2 & 32.1 & 63.2 & 69.8 & 53.4 & 61.9 & 65.9\\
\cmidrule(lr){3-13}
 &  & StreamingLLM & 26.0 & 15.3 & 28.7 & 79.0 & 30.3 & 54.7 & 65.5 & 46.6 & 55.2 & 44.6\\
 &  & FastV & 69.3 & 28.7 & 76.7 & 81.1 & 31.2 & 58.7 & 67.0 & 50.1 & 58.0 & 57.9\\
 &  & PruMerge & 83.3 & 36.0 & 83.3 & 79.4 & 30.8 & 60.0 & 68.6 & 50.7 & 56.0 & 60.9\\
 &  & Ours & \textbf{90.0} & \textbf{48.7} & \textbf{87.3} & \textbf{81.8} & \textbf{31.7} & \textbf{61.3} & \textbf{69.9} & \textbf{53.0} & \textbf{59.7} & \textbf{64.8}\\
\cmidrule(lr){2-13}
 & \multirow{5}{*}{72B}
 & Original & 89.3 & 66.0 & 88.0 & 85.3 & 32.3 & 70.9 & 77.3 & 65.0 & 63.9 & 70.9\\
 \cmidrule(lr){3-13}
 &  & StreamingLLM & 33.3 & 20.0 & 35.3 & 81.9 & 30.6 & 62.6 & 72.9 & 60.2 & 58.0 & 50.5\\
 &  & FastV & 22.0 & 48.7 & 77.3 & 83.7 & 31.5 & 65.9 & 73.7 & 62.6 & 61.7 & 58.6\\
 &  & PruMerge & 85.3 & 58.0 & 86.0 & 82.0 & 31.4 & 66.7 & 74.8 & 62.6 & 58.6 & 67.3\\
 &  & Ours & \textbf{90.0} & \textbf{63.3} & \textbf{88.0} & \textbf{84.6} & \textbf{32.0} & \textbf{69.0} & \textbf{76.7} & \textbf{63.2} & \textbf{63.0} & \textbf{70.0}\\
\midrule
\multirow{16.5}{*}{NVILA}
 & \multirow{5}{*}{2B}
 & Original & 90.0 & 22.0 & 87.3 & 71.2 & 6.6 & 50.9 & 53.2 & 42.3 & 50.7 & 52.7\\
 \cmidrule(lr){3-13}
 &  & StreamingLLM & 26.0 & 12.7 & 34.7 & 69.0 & 5.8 & 45.7 & 50.1 & 40.7 & 49.1 & 37.1\\
 &  & FastV & 50.7 & 14.7 & 56.7 & 70.7 & 7.2 & 46.7 & 50.6 & 41.1 & \textbf{50.1} & 43.2\\
 &  & PruMerge & 27.3 & \textbf{31.3} & 81.3 & 67.7 & 11.1 & 47.3 & 50.4 & 42.2 & 48.0 & 45.2\\
 &  & Ours & \textbf{89.3} & 27.3 & \textbf{87.3} & \textbf{71.8} & \textbf{20.1} & \textbf{50.4} & \textbf{53.1} & \textbf{45.2} & 49.5 & \textbf{54.9}\\
\cmidrule(lr){2-13}
 & \multirow{5}{*}{8B}
 & Original & 98.7 & 40.7 & 100.0 & 81.7 & 33.0 & 63.9 & 68.3 & 52.0 & 67.5 & 67.3\\
 \cmidrule(lr){3-13}
 &  & StreamingLLM & 30.0 & 17.3 & 41.3 & 78.4 & 30.8 & 54.3 & 63.7 & 46.2 & 58.1 & 46.7\\
 &  & FastV & 87.3 & 33.3 & 90.7 & 80.4 & 32.5 & 59.5 & 66.8 & 50.5 & 64.5 & 62.8\\
 &  & PruMerge & 4.7 & 32.0 & 93.3 & 77.1 & 31.4 & 56.9 & 65.1 & 49.4 & 57.9 & 52.0\\
 &  & Ours & \textbf{96.0} & \textbf{38.0} & \textbf{98.7} & \textbf{80.7} & \textbf{32.5} & \textbf{61.1} & \textbf{68.2} & \textbf{52.5} & \textbf{65.0} & \textbf{65.9}\\
\cmidrule(lr){2-13}
 & \multirow{5}{*}{15B}
 & Original & 95.3 & 42.0 & 100.0 & 78.7 & 30.9 & 65.8 & 72.3 & 58.2 & 60.5 & 67.1\\
 \cmidrule(lr){3-13}
 &  & StreamingLLM & 34.0 & 18.7 & 34.0 & 74.0 & 28.5 & 58.5 & 65.1 & 53.7 & 55.0 & 46.8\\
 &  & FastV & 48.7 & 24.7 & 80.7 & 77.0 & 30.6 & 60.6 & 69.1 & 56.7 & 57.3 & 56.2\\
 &  & PruMerge & 19.3 & 43.3 & 98.0 & 72.4 & 30.0 & 59.3 & 68.4 & 52.3 & 52.8 & 55.1\\
 &  & Ours & \textbf{94.0} & \textbf{52.7} & \textbf{99.3} & \textbf{77.7} & \textbf{31.2} & \textbf{63.5} & \textbf{70.8} & \textbf{57.8} & \textbf{58.4} & \textbf{67.3}\\
\midrule
\multirow{4}{*}{MiniCPM-V}
 & \multirow{4}{*}{8B}
 & Original & 88.7 & 36.7 & 88.7 & 78.9 & 13.8 & 58.5 & 60.3 & 53.4 & 55.0 & 59.3\\
 \cmidrule(lr){3-13}
 &  & StreamingLLM & 22.0 & 15.3 & 28.7 & 76.0 & \textbf{23.2} & 53.8 & 56.7 & 48.2 & 51.3 & 41.7\\
 &  & FastV & 82.7 & 26.7 & 71.3 & 78.0 & 14.8 & 56.7 & 58.2 & 51.8 & 53.2 & 54.8\\
 &  & Ours & \textbf{89.3} & \textbf{41.3} & \textbf{89.3} & \textbf{78.2} & 16.3 & \textbf{57.4} & \textbf{59.5} & \textbf{52.3} & \textbf{53.6} & \textbf{59.7}\\
\bottomrule
\end{tabular}
\caption{Performance comparison across different model families, sizes, and methods on five benchmarks at a 30\% relative token budget.}
\label{tab:overall_performance}
\end{table*}

\xhdr{Overall Performance}
\name consistently outperforms state-of-the-art token compression methods across multiple model families, sizes, and benchmarks, matching dense model performance at a 30\% token budget (Table~\ref{tab:overall_performance}). \name maintains a maximum relative average performance drop of just 2.4\% across six models, whereas StreamingLLM, FastV, and PruMerge incur drops of 35.8\%, 19.7\%, and 15.4\%, respectively. On VideoNIAH benchmark, \name shows a maximum relative drop of only 2.8\%, compared to 38.7-69.5\% for other methods. VideoMME is sensitive because it relies on spatial and temporal details, which cannot be answered by LLM common sense alone.
Excluding VideoNIAH, \name's maximum drop is 3.6\%, significantly lower than the 8.3-14.2\% seen in baselines. Detailed statistics are in Appendix~\ref{sec:appendix/performance}.

\begin{figure}
    \centering
    \includegraphics[width=\linewidth]{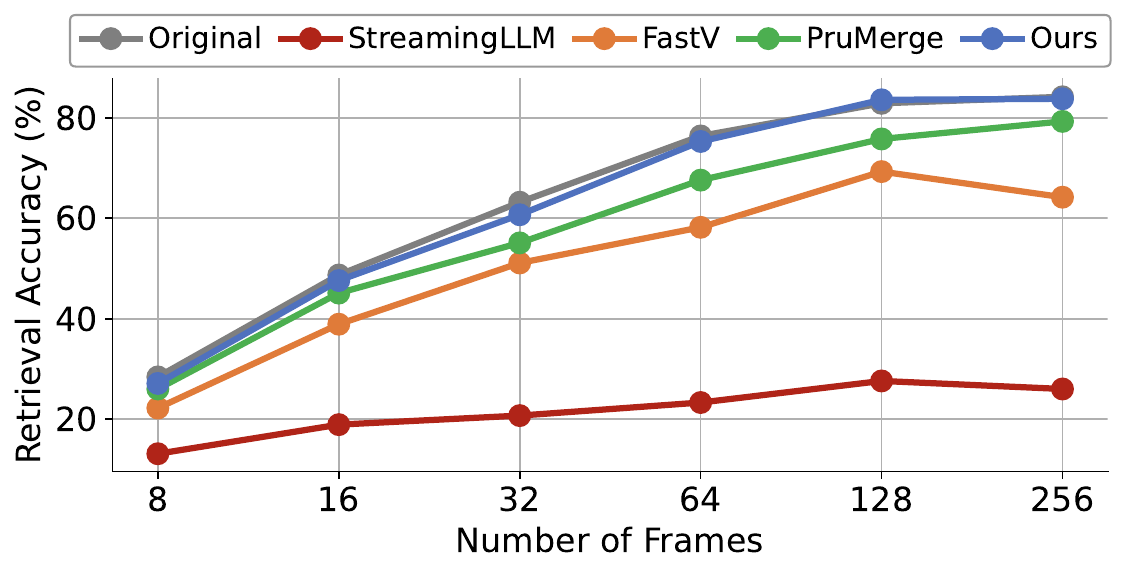}
    \vspace{-15pt}
    \caption{The VideoNIAH performances for the Llava-Video-7B across various numbers of input frames.}
    \label{fig:input_length}
    \vspace{-12pt}
\end{figure}

\xhdr{Scaling Input Length}
Figure~\ref{fig:input_length} investigates model performance across varying frame counts, from 8 to 256 frames. \name consistently outperforms baseline methods, achieving retrieval accuracies close to the original dense model. As the frame count increases, the performance gap between \name and the dense model shrinks from 4.6\% to 0.5\%, significantly outperforming the best baseline, which has gaps ranging from 12.8\% down to 5.8\%.
We also explore the scalability of \name across different token budgets, models and benchmarks in Appendix \ref{sec:appendix/performance}.

\subsection{Efficiency}


\begin{figure}
    \centering
    \includegraphics[width=\linewidth]{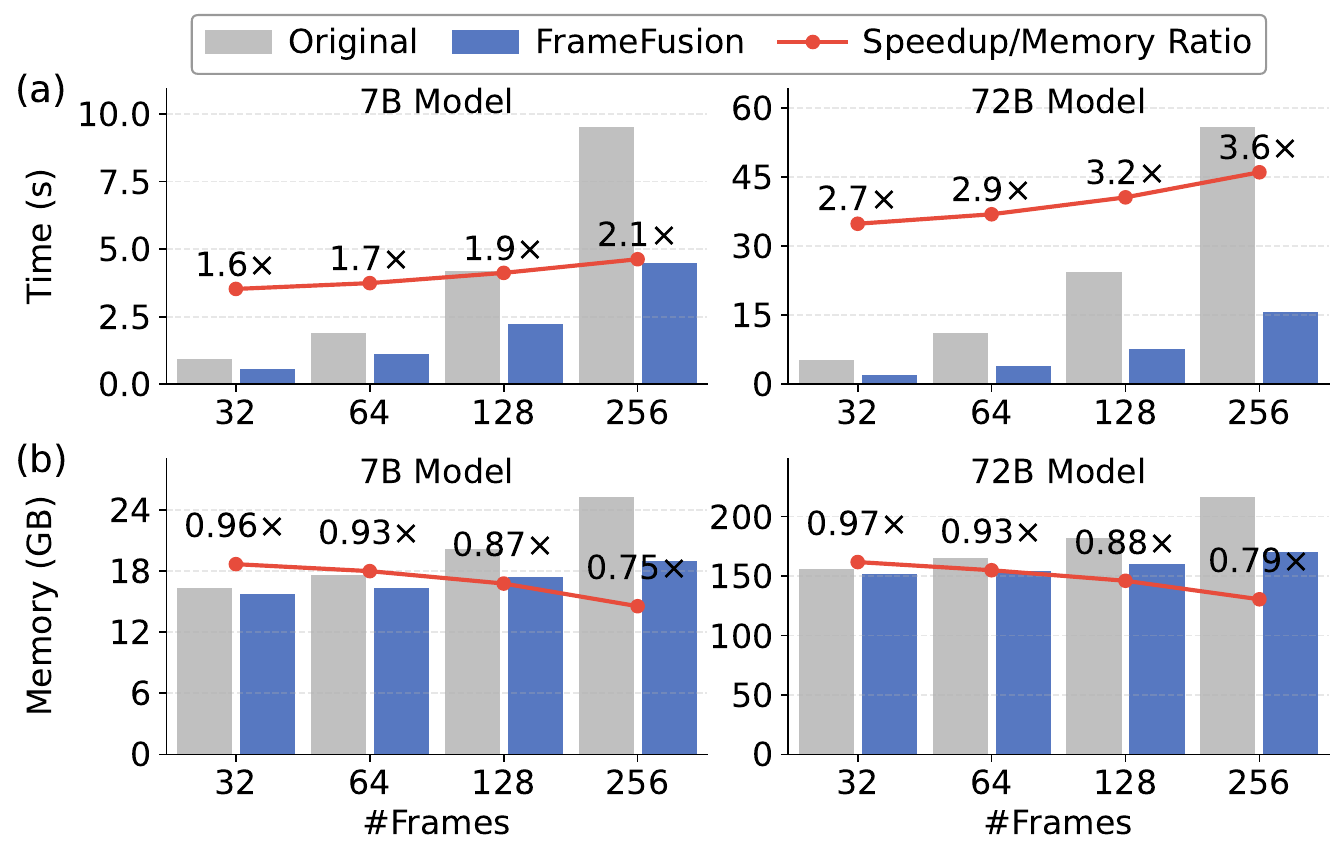}
    \vspace{-20pt}
    \caption{End-to-end runtime and memory consumption of Llava-Video-7B and 72B models with FrameFusion, using one and four NVIDIA A100-80GB GPUs for 7B and 72B models, respectively.}
    \label{fig:speedup_7b_72b}
\end{figure}

We evaluate the wall-clock speedup and GPU memory reduction of \name in Figure~\ref{fig:speedup_7b_72b}, across 7 to 72B model sizes and 32 to 256 frames. Results are averaged over 128 videos from the VideoMME benchmark. 
Additional results, including runtime and memory breakdowns, evaluations on more models and token budgets, and token reduction details, are provided in Appendix~\ref{sec:appendix/efficiency}. Asymptotic complexity analysis are in Appendix~\ref{sec:appendix/asymptotic_complexity}.

\xhdr{Time}
\name achieves end-to-end speedups of 1.6-3.6$\times$ with a 30\% token budget. 
Compression stages in \name account for only 0.6-4\% of the total runtime.
The speedup grows as the number of frames increases, owing to the reduced $O((CN)^2)$ complexity in attention computations.
Since \name speedups the LLM computation in LVLMs, it yields greater speedups with the larger 72B model, where overhead such as video sampling and ViT encoding becomes proportionally less significant. 

\xhdr{Memory} 
\name reduces GPU memory for KV-Cache and activations by the factor of token budget. As the number of frames grows and the dominance of model parameters decreases, memory savings continue to improve.


\subsection{Ablation Study}
\label{sec:ablation}

We validate the impact of each observation and design choice in \name, showing their individual effectiveness with average scores on VideoNIAH, VideoMME (without subtitles), and NExT-QA benchmarks.
We further analyze the effects of similarity computation strategies, distance metrics, similarity thresholds, and positional embeddings, as detailed in Appendix~\ref{sec:appendix/ablation}.


\xhdr{Design Choice~\ref{dsn:corr_token_only}}
Instead of computing $N \times N$ cosine similarities with significant overhead, \name calculates only $N$ token similarities between corresponding visual tokens in adjacent frames.
We compare our approach with two common alternative strategies:
\textit{1. Adj. token}, which computes $N$ similarities between adjacent image patches (i.e., adjacent visual tokens).
\textit{2. Random}, which calculates similarities for $N$ randomly selected token pairs.
We compare the three similarity calculation strategies on three benchmarks at a 30\% token budget. As shown in Table~\ref{tab:merging_strategy}, our strategy achieves 11\% and 8\% higher average accuracies than the other two strategies.
We also examine the choice of similarity calculation metrics by comparing cosine similarity with alternative distance metrics, including the inner product, Minkowski-2, and Minkowski-1 distance. Experiments show that cosine similarity yields 2-5\% average score gains over the alternative metrics. The detailed numbers are presented in Appendix~\ref{sec:distance_metrices}. 

\begin{table}[tb]
    \centering
    \setlength\tabcolsep{2pt} 
    \begin{tabular}{l|ccc|c}
    \toprule
    \textbf{Design~\ref{dsn:corr_token_only}}& \textbf{VideoNIAH} & \textbf{VideoMME} & \textbf{NExT-QA} & \textbf{Avg.}\\
    \midrule
    Random  & 60.0 & 59.0 & 56.6 &58.5 \\
    Adj. token   & 64.0 & 59.9 & 56.6 & 60.2 \\
    Ours   & \textbf{76.5}  & \textbf{61.3} & \textbf{56.8} & \textbf{64.9} \\
    \bottomrule
    \end{tabular}
    \vspace{-8pt}
    \caption{Performance of different similarity calculation strategies with the same relative token budget of 30\% on VideoNIAH, VideoMME, and NExT-QA.}
    \label{tab:merging_strategy}
\end{table}


\begin{table}[tb]
    \centering
    \setlength\tabcolsep{2pt} 
    \begin{tabular}{lc|ccc|c}
    \toprule
    \textbf{Layer} & \textbf{Rate} & \textbf{VideoNIAH}  & \textbf{VideoMME} &\textbf{NExT-QA} & \textbf{Avg}\\
    \midrule
    \multicolumn{2}{c|}{Original} & 76.4  & 63.2 &57.7 & 65.8\\
    \midrule
    0   & 50.0\%    & 76.2  & 62.7 &57.4  & 65.4\\
    1   & 52.0\%      & 76.8  & 62.6 &57.4  & 65.6\\
    2   & 53.8\%    & 76.4  & 62.0 &57.3  & 65.2\\
    12  & 87.5\%    & 74.4  & 60.3 &56.4 & 63.7\\
    13  & 93.3\%   & 64.2  & 57.9 &55.6 & 59.2\\
    14  & 100.0\%   & 48.9  & 52.6 &50.9 & 50.8\\
    \bottomrule
    \end{tabular}
    \vspace{-8pt}
    \caption{Performance of Llava-Video-7B with cascaded token merging at different layers, where merging rates are adjusted to maintain an average token count (token budget) of 50\% across all layers.}
    \vspace{-12pt}
    \label{tab:merging_layer}
\end{table}


\begin{table}[tb]
    \centering
    \setlength\tabcolsep{2pt} 
    \begin{tabular}{l|ccc|c}
        \toprule
        \textbf{Design~\ref{dsn:merge_then_prune}} &  \textbf{VideoNIAH} &\textbf{VideoMME} & \textbf{NExT-QA}  & \textbf{Avg}\\ 
        \midrule
        Prune $\rightarrow$ merge&  73.1 &59.9 & 55.9 & 63.0\\
        Merge $\rightarrow$ prune&  \textbf{73.3} &\textbf{60.9} & \textbf{56.6}  & \textbf{63.6}\\ 
        \bottomrule
    \end{tabular}
    \vspace{-8pt}
    \caption{Performance of the Llava-Video-7B model with different orders of merging and pruning.}
    \label{tab:order}
\end{table}

\xhdr{Design Choice~\ref{dsn:merge_then_prune}}
\name first merges at the initial layers, then prunes at subsequent layers. 
We evaluate the influence of the merging layer position. 
As shown in Table~\ref{tab:merging_layer}, delaying the merging to later layers requires a higher merging rate to meet the given relative token budget of 50\%. 
The reduced high similarity values at deeper layers also causes significant performance drops compared with merging at shallower layers.
Additionally, we examine the effect of the order of merging and pruning. As shown in Table~\ref{tab:order}, with a fixed token budget of 30\% and the same number of tokens reduced in layers 1 and 2, merging before pruning achieves better performance compared to pruning before merging.


\begin{table}[tb]
    \centering
    \setlength\tabcolsep{2pt} 
    \begin{tabular}{l|ccc|c}
        \toprule
        \textbf{Design~\ref{dsn:cascaded_merging}} &  \textbf{VideoNIAH} &\textbf{VideoMME} & \textbf{NExT-QA}  & \textbf{Avg}\\ 
        \midrule
        Non-cascaded & 74.9 & 60.0 & 55.4 & 63.4\\
        Cascaded & \textbf{76.0} & \textbf{62.8} & \textbf{57.5} & \textbf{65.4}\\
        \bottomrule
    \end{tabular}
    \vspace{-6pt}
    \caption{Performance of the Llava-Video-7B model with different orders of merging and pruning.}
    \label{tab:cascaded}
\end{table}

\xhdr{Design Choice~\ref{dsn:cascaded_merging}}
\name adopts a cascaded merging strategy, where merged tokens remain reduced across layers to maximize efficiency.
To evaluate the accuracy-efficiency trade-offs of cascaded merging, we also implement non-cascaded merging for comparison. 
Following previous works~\citep{tang2024quest, li2024snapkv, Zhang2023H2O}, this baseline performs the exact same merging strategy, but only on Key Value metrics in the attention module to avoid prominently removing any tokens. As a result, it leaves FFN computations unchanged.
Its merging rate across all layers are set to be 30\%. Our cascaded counterpart uses the same computation FLOPs, translating to an 84\% relative token budget.
As shown in Table~\ref{tab:cascaded}, under the same FLOPs, \name outperforms the non-cascaded counterparts by 3\% in average score. showing comparable performance to the original model.
\section{Conclusion}

In this paper, we propose \name, a similarity-based token merging method for video LVLMs. By combining similarity-based merging with importance-based pruning, \name reduces redundant visual tokens while retaining critical information. This approach optimizes computational efficiency and memory usage, enabling accurate video understanding with significantly fewer tokens. Experiments across multiple benchmarks demonstrate a 70\% reduction in visual tokens with minimal performance loss, achieving 1.6-3.6$\times$ end-to-end speedups. \name providing new insights in token similarity for LVLMs, offering an efficient and scalable solution for real-world video language applications.

\clearpage
\section*{Acknowledgement}
This work was supported by National Natural Science Foundation of China (No. 62325405, 62104128, U19B2019, U21B2031, 61832007, 62204164, 92364201), Tsinghua EE Xilinx AI Research Fund, and Beijing National Research Center for Information Science and Technology (BNRist). We thank all the support
from Infinigence-AI.
{
    \small
    \bibliographystyle{ieeenat_fullname}
    \bibliography{main}
}

\clearpage
\setcounter{page}{1}
\maketitlesupplementary

\section{Detailed Experiment Setup}

\label{sec:appendix/experiment_setup}
\subsection{Method Setup}
\subsubsection{Baseline Setup}
For the baselines StreamingLLM~\citep{xiao2023streamingLLM} and FastV~\citep{chen2024fastv}, we follow the official implementations and set the attention sink size of StreamingLLM to 8 and $K$ in FastV to 2.

\subsubsection{\name Setup} 

\begin{figure*}[ht]
\centering
\includegraphics[width=\linewidth]{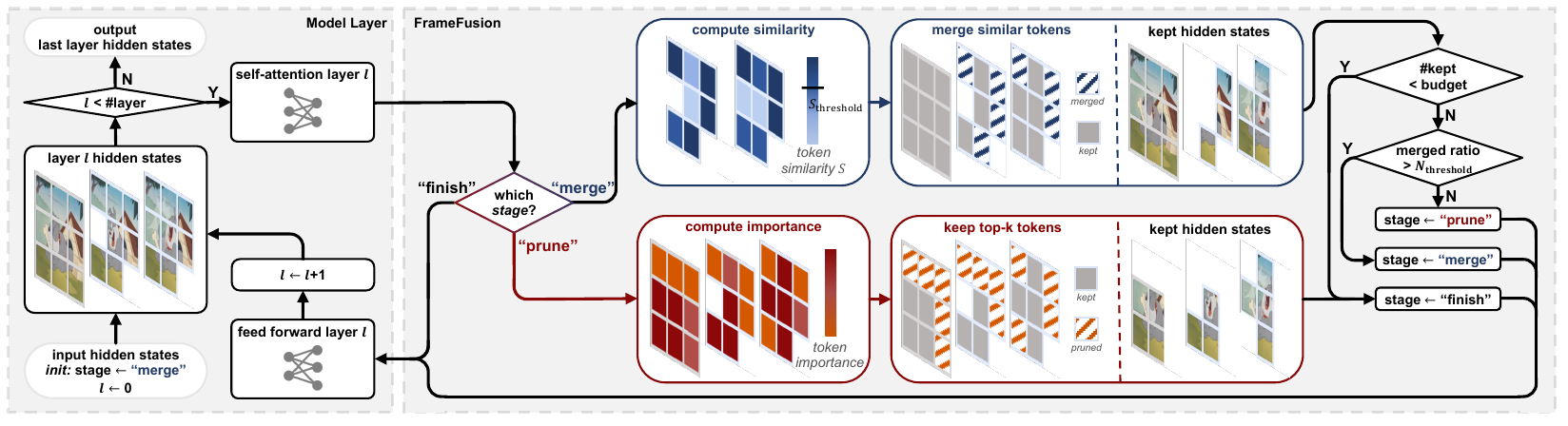}
\caption{The workflow of \name when applied to LVLMs. At each layer, \name performs merging, pruning, or no action between the self-attention and feed-forward layers, depending on the current stage. The stage initially starts as ``merge'' and updates according to transition conditions.}
\vspace{8pt}
\label{fig:diagram}
\end{figure*}

\xhdr{Workflow details} 
For \name, token merging is only applied to visual tokens because they dominate input length and show higher similarity between adjacent frames, enabling $O(N)$ complexity merging.
The detailed workflow of \name is shown in Figure \ref{fig:diagram}.

\xhdr{Hyperparameters} The merging ratios across layers are controlled by two hyperparameters: $S_{\text{threshold}}$ and $N_{\text{threshold}}$, as discussed in Section~\ref{sec:method/design}.

$S_{\text{threshold}}$ defines the minimum cosine similarity required for two tokens to be considered similar and merged. 
Since similarity distributions vary across models, we set $S_{\text{threshold}}$ to match the median of similarity at the first model layer under typical input cases, such as 128 samples from the VideoMME dataset. 
For the Llava-Video series, we set $S_{\text{threshold}} = 0.6$; for MiniCPM-V, we set $S_{\text{threshold}} = 0.7$; for NVILA-2B,8B,15B, We set $S_{\text{threshold}} =0.6, 0.75, 0.8$, respectively.

$N_{\text{threshold}}$ determines the transition from merging to pruning. If the number of similar tokens (tokens with cosine similarity above $S_{\text{threshold}}$) falls below $N_{\text{threshold}}$, the model switches to pruning. We set $N_{\text{threshold}} = 0.1$ to avoid extensive similarity computations across the entire model.

To ensure the merging process does not excessively reduce the token count below the predefined token budget $C$, we precompute the maximum number of token pairs ($N_{\text{max}}$) that can be merged per layer. If the actual number of pairs exceeds $N_{\text{max}}$, only the top $N_{\text{max}}$ pairs with the highest cosine similarity are merged. Any remaining merging or pruning steps are skipped, and the model proceeds with a standard forward pass.

\subsection{Model Setup}
We follow the default frame count settings for all models, except for NVILA-Lite-2B. Since NVILA-Lite-2B is not specifically trained for video tasks, we set its frame count to 64. For the Llava-Video series and Minicpm-V, the frame count is set to 64, while for NVILA-Video-8B and NVILA-Video-15B, it is set to 256.

\section{Additional Experiment Results}
\subsection{Performance}
\label{sec:appendix/performance}

\subsubsection{Computation-Accuracy Trade-off}
\label{sec:appendix/trade-off}


\begin{figure}[tb]
    \centering
    \begin{subfigure}[b]{\linewidth}
        \centering
        \includegraphics[width=\linewidth]{fig/pareto_label.pdf}
        \vspace{-14pt}
    \end{subfigure}

    \begin{subfigure}[b]{\linewidth}
        \centering
        \includegraphics[width=\linewidth]{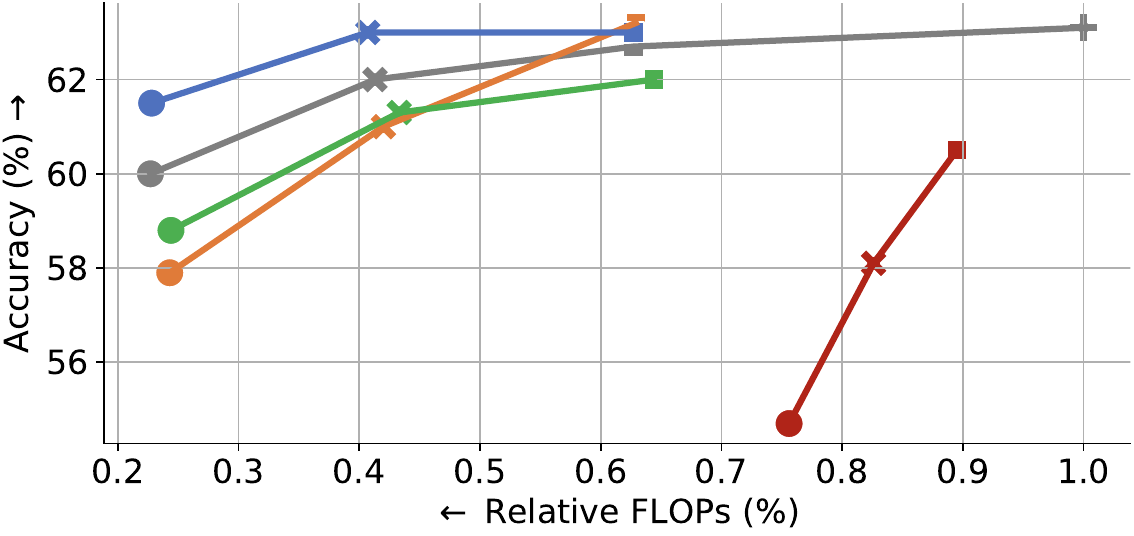}
        \vspace{-8pt}
    \end{subfigure}

    \caption{The accuracy-computation trade-offs of various token compression methods, tested on Llava-Video-7B with VideoMME benchmark. Original* represents the original model with reduced frame rates.}
    \label{fig:pareto_VideoMME}
    \vspace{5pt}
\end{figure}


\begin{figure}[tb]
    \centering
    \begin{subfigure}[b]{\linewidth}
        \centering
        \includegraphics[width=\linewidth]{fig/pareto_label.pdf}
        \vspace{-14pt}
    \end{subfigure}

    \begin{subfigure}[b]{\linewidth}
        \centering
        \includegraphics[width=\linewidth]{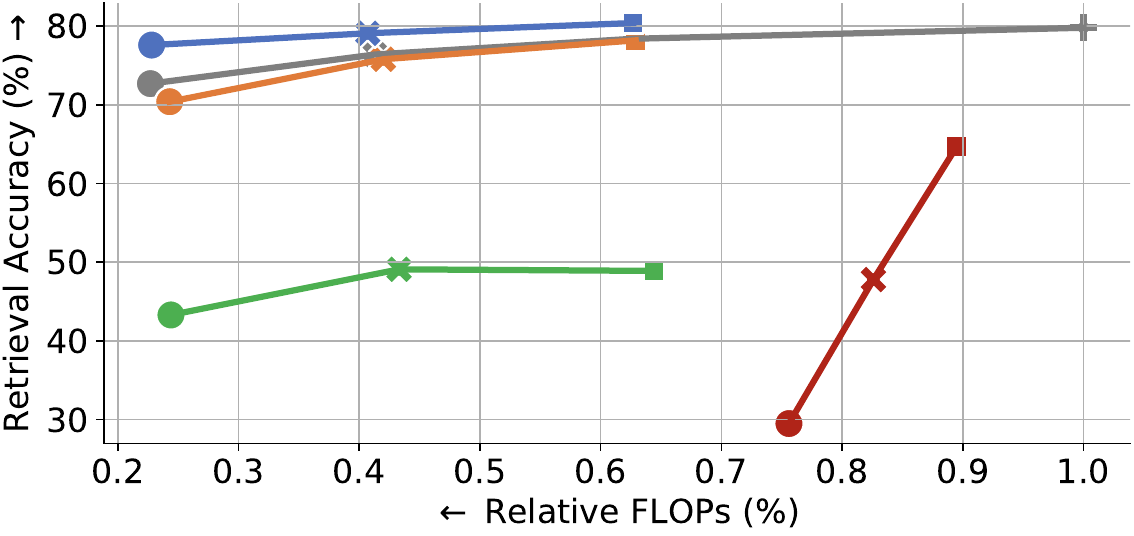}
        \vspace{-8pt}
    \end{subfigure}

    \caption{The accuracy-computation trade-offs of various token compression methods, tested on NVILA-8B with VideoNIAH benchmark. Original* represents the original model with reduced frame rates.}
    \label{fig:pareto_niah}
\end{figure}

We further investigate the trade-off between computational cost and accuracy. We evaluate the Llava-Video-7B and NVILA-8B models on the VideoMME and VideoNIAH benchmark, respectively. The results are shown in Figure ~\ref{fig:pareto_VideoMME} and Figure ~\ref{fig:pareto_niah}. As the number of FLOPs decreases, other baseline methods exhibit a noticeable decline in accuracy, whereas \name maintains superior performance.

\subsubsection{Performance Across Different Input Length}

\begin{figure}
    \centering
    \includegraphics[width=\linewidth]{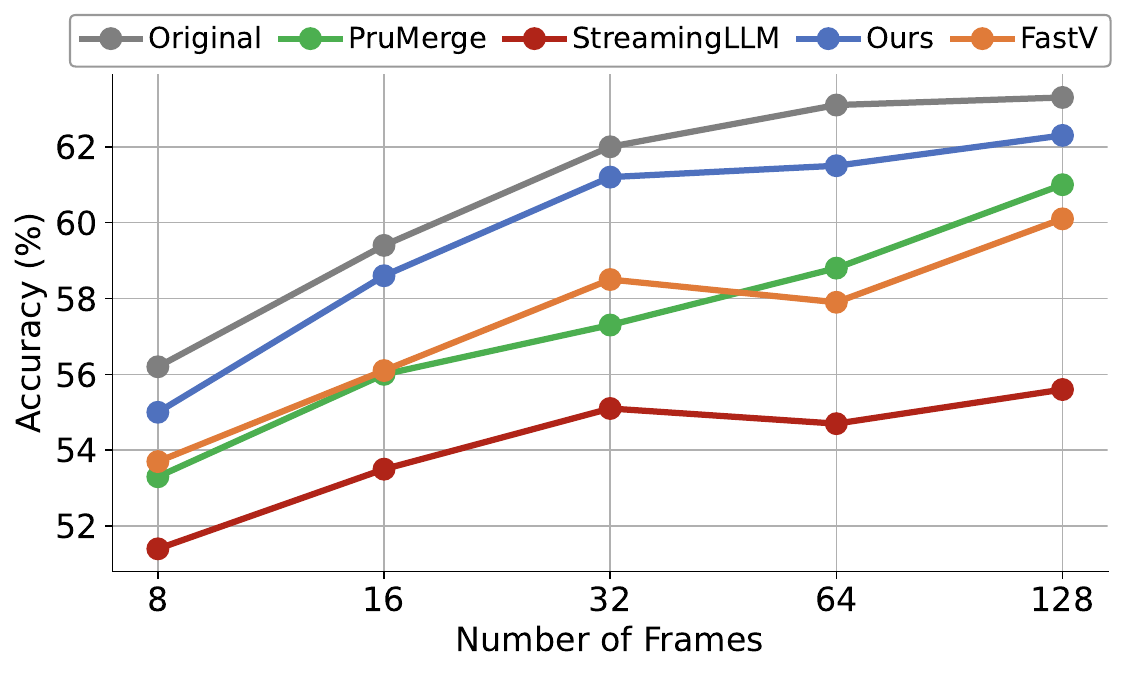}
    \caption{The VideoMME performances for the Llava-Video-7B across various numbers of input frames.}
    \label{fig:input_length_VideoMME}
\end{figure}

Figure~\ref{fig:input_length_VideoMME} presents the performance of Llava-Video-7B on the VideoMME benchmark as the number of input frames varies from 8 to 128. Across all configurations, \name consistently outperforms the baseline methods, demonstrating its robustness to different input length.

\subsubsection{Performance Across Different Token Budgets}
\label{sec:performance_across_budget}

\begin{table*}[t]
    \centering
    \begin{tabular}{llc|cc|cc|cc|c}
        \toprule
        ~ & ~ & ~ & \multicolumn{2}{c|}{VideoMME} & \multicolumn{2}{c|}{NExt-QA-MC} & \multicolumn{2}{c|}{NExt-QA-OE} & ~ \\
        Model & Method & Budget & Score $\uparrow$ & Drop $\downarrow$ & Score $\uparrow$ & Drop $\downarrow$ & Score $\uparrow$ & Drop $\downarrow$ & Max. Drop $\downarrow$ \\
        \midrule
        \multirow{4}{*}{Llava-Video-7B} 
        & Original & 1.0 & 63.2 & - & 83.2 & - & 32.1 & - & - \\
        \cmidrule(lr){2-10}
        & \multirow{3}{*}{Ours} 
        & 0.3 & 61.3 & 3.0\% & 81.8 & 1.7\% & 31.7 & 1.2\% & 3.0\% \\
        & & 0.5 & 62.6 & 0.9\% & 82.7 & 0.6\% & 32.1 & 0.0\% & 0.9\% \\
        & & 0.7 & 63.0 & 0.3\% & 82.8& 0.5\%& 32.1& 0.0\%& 0.5\%\\
        \midrule
        \multirow{4}{*}{MiniCPM-V-8B} 
        & Original & 1.0 & 58.5 & - & 78.9 & - & 13.8 & - & - \\
        \cmidrule(lr){2-10}
        & \multirow{3}{*}{Ours} 
        & 0.3 & 57.4 & 1.9\% & 78.2 & 0.9\% & 16.3 & -18.1\% & 1.9\% \\
        & & 0.5 & 58.5 & 0.0\% & 78.6 & 0.4\% & 17.4 & -26.1\% & 0.4\% \\
        & & 0.7 & 57.8 & 1.2\% & 78.6 & 0.4\% & 16.1 & -16.7\% & 1.2\% \\
        \bottomrule
    \end{tabular}
    \caption{Performance comparison between the original and proposed methods on VideoMME, NExt-QA-MC, and NExt-QA-OE benchmarks with different relative token budgets on Llava-Video-7B model. Drop indicates the relative performance decrease compared to the original method.}
    \vspace{8pt}
    \label{tab:performance_comparison_drop}
\end{table*}

Table~\ref{tab:performance_comparison_drop} presents the benchmark performance of the Llava-Video-7B model at token budgets ranging from 0.3 to 0.7. At a 30\% token budget, \name achieves strong performance, with a maximum relative drop of less than 3.0\% compared to the dense model. As the budget increases to 0.5 and 0.7, the maximum drops further decrease to $\leq$1.2\%.

\subsubsection{Performance Across Different Models}
\label{sec:appendix/experiment/numeric}

We present the detailed numeric results of the scalability experiments in Section~\ref{sec:exp/performance}.

\begin{figure*}
    \centering
    \includegraphics[width=\linewidth]{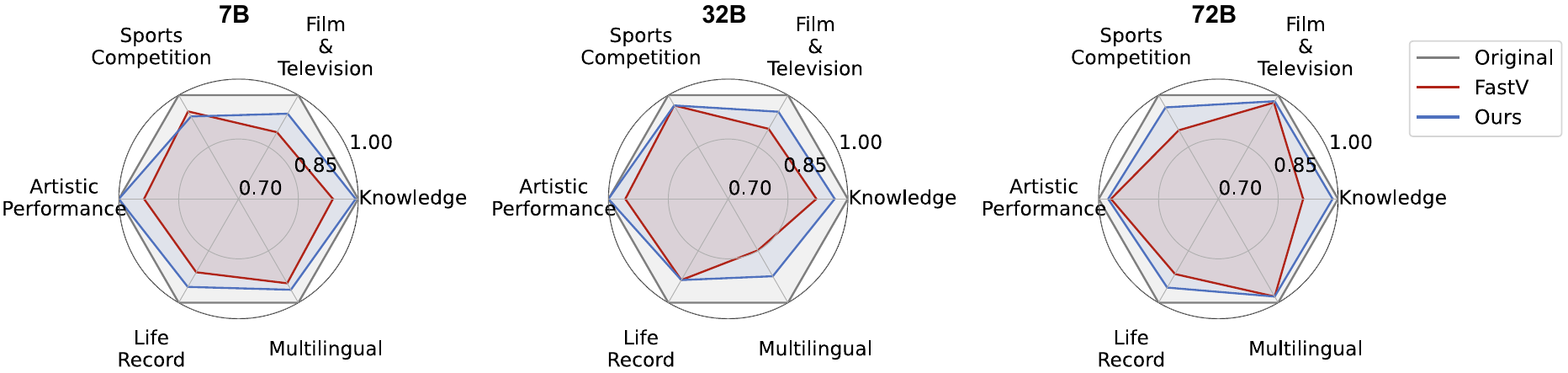}
    \caption{The VideoMME performance for each category across Llava-Video-7B, 32B, and 72B for different methods. All scores are normalized by the original model.}
    \label{fig:scaling_model_size}
\end{figure*}

\begin{table*}[tb]
    \setlength\tabcolsep{6pt}
    \centering
    \begin{tabular}{ll|ccc|cccccc}
    \toprule
    \textbf{Model} & \textbf{Method} & 
       \textbf{Short}&\textbf{Medium}&\textbf{Long}&\makecell{\textbf{KL}} &
    \makecell{\textbf{FT}} & \makecell{\textbf{SC}} & \makecell{\textbf{AP}} & \makecell{\textbf{LR}} & \makecell{\textbf{ML}}  \\
    \midrule
    \multirow{5}{*}{Llava-Video-7B} 
    & Original & 75.8 & 61.7 & 52.2 & 63.1 & 67.2 & 61.8 & 61.7 & 63.7 & 58.9 \\
     \cmidrule(lr){2-11}
    & StreamingLLM& 63.4 & 54.1 & 46.4 & 55.1& 57.2& 56.0& 54.2& 52.9&48.9\\
    & FastV & 68.4 & 58.0 & 49.6 & 59.1 & 60.0 & 58.9 & 57.8 & 58.1 & 55.6 \\
    & PruMerge & 69.7 & \textbf{60.1} & \textbf{50.2} & 59.1 & 63.6 & \textbf{59.1} & 58.9 & 60.6 & \textbf{57.8} \\
    & Ours & \textbf{74.0} & 59.8 & 50.0 & \textbf{62.7} & \textbf{63.6} & 58.0 & \textbf{61.7} & \textbf{60.8} & 56.7 \\
    \midrule
    \multirow{5}{*}{Llava-Video-72B} 
    & Original & 80.9 & 69.7 & 62.1 & 73.2 & 74.4 & 68.0 & 71.4 & 68.9 & 62.2 \\
    \cmidrule(lr){2-11}
    & StreamingLLM & 68.2 & 59.9 &59.8  & 65.7 & 66.7 & 59.3 & 65.6 & 58.7 & 58.9 \\
    & FastV & 73.0 & 64.9 & 60.2 & 66.8 & 72.8 & 61.1 & 69.2 & 63.2 & 61.1 \\
    & PruMerge & 74.0 & 65.8 & 60.3 & 70.4 & \textbf{73.6} & 62.9 & 68.3 & 61.6 & 54.4 \\
    & Ours & \textbf{78.3} & \textbf{67.9} & \textbf{60.9} & \textbf{72.2} & 73.1 & \textbf{65.6} & \textbf{69.7} & \textbf{65.9} & \textbf{61.1} \\
    \midrule
    \multirow{5}{*}{NVILA-2B} 
    & Original & 61.4 & 48.9 & 42.4 & 47.2 & 56.4 & 49.8 & 55.0 & 51.0 & 52.2 \\
    \cmidrule(lr){2-11}
    & StreamingLLM & 52.9 & 43.9 & 40.3 & 43.0 & 49.2 & 46.2 & 50.6 & 44.4 & 43.3 \\
    & FastV & 53.7 & 45.6 & 40.8 & 43.8 & 49.7 & 46.2 & 51.4 & 46.7 & 43.3 \\
    & PruMerge & 53.9 & 45.0 & \textbf{43.1} & 43.5 & 52.8 & 45.8 & 51.7 & 48.1 & 45.6 \\
    & Ours & \textbf{61.3} & \textbf{47.0} & 43.0 & \textbf{48.3} & \textbf{55.6} & \textbf{48.2} & \textbf{55.3} & \textbf{49.8} & \textbf{45.6} \\
    \midrule
    \multirow{5}{*}{NVILA-8B} 
    & Original & 74.9 & 62.1 & 54.7 & 64.8 & 66.4 & 62.2 & 61.9 & 63.7 & 63.3 \\
    \cmidrule(lr){2-11}
    & StreamingLLM & 61.2 & 53.8 & 48.0 & 54.9 & 57.8 & 54.7 & 52.5 & 52.2 & 55.6 \\
    & FastV & 72.0 & 56.7 & 50.0 & 60.7 & 62.8 & 57.6 & 57.8 & 58.9 & 57.8 \\
    & PruMerge & 67.6 & 54.9 & 48.3 & 57.3 & 61.1 & 56.0 & 54.7 & 56.2 & 55.6 \\
    & Ours & \textbf{74.2} & \textbf{57.7} & \textbf{51.3} & \textbf{60.7} & \textbf{65.3} & \textbf{59.3} & \textbf{58.6} & \textbf{62.1} & \textbf{58.9} \\
    \midrule
    \multirow{5}{*}{NVILA-15B} 
    & Original & 77.3 & 64.7 & 55.3 & 67.2 & 68.1 & 62.7 & 63.3 & 66.2 & 66.7 \\
    \cmidrule(lr){2-11}
    & StreamingLLM & 63.8 & 57.4 & 54.3 & 60.6 & 60.6 & 55.6 & 58.6 & 56.5 & 60.0 \\
    & FastV & 69.2 & 58.7 & 53.9 & 62.8 & 63.3 & 57.1 & 60.8 & 58.1 & 63.3 \\
    & PruMerge & 66.0 & 59.3 & 52.6 & 61.0 & 61.1 & 55.1 & 57.5 & 59.8 & 61.1 \\
    & Ours & \textbf{73.2} & \textbf{62.3} & \textbf{55.0} & \textbf{64.6} & \textbf{68.1} & \textbf{60.9} & \textbf{61.1} & \textbf{62.5} & \textbf{65.6} \\
    \midrule
    \multirow{5}{*}{MiniCPM-V-8B} 
    & Original & 69.1 & 56.6 & 49.8 & 59.0 & 63.6 & 54.2 & 63.3 & 54.9 & 60.0 \\
    \cmidrule(lr){2-11}
    & StreamingLLM & 61.1 & 51.8 & 48.4 & 54.6 & 58.1 & 52.2 & 56.4 & 49.7 & 55.6 \\
    & FastV & 67.1 & 53.9 & \textbf{49.2} & 57.2 & 59.2 & 53.8 & \textbf{60.8} & \textbf{54.6} & 56.7 \\
    & Ours & \textbf{69.7} & \textbf{54.1} & 48.3 & \textbf{57.9} & \textbf{63.1} & \textbf{53.8} & 60.3 & 54.4 & \textbf{56.7} \\
    \bottomrule
    \end{tabular}
    \caption{Numeric VideoMME scores of different methods and model sizes across various video categories. ``KL'', ``FT'', ``SC'', ``AP'', ``LR'', ``ML'' are short for ``Knowledge'', ``Film \& Television'', ``Sports Competition'', ``Artistic Performance'', ``Life Record'', and ``Multilingual''.}
    \label{tab:model_size_detail}
\end{table*}

As shown in Figure~\ref{fig:scaling_model_size}, \name consistently outperforms FastV baseline across all model sizes and VideoMME categories, demonstrating comparative performance with the original model at a 30\% relative token budget.
Note that the model Llava-Video-32B has been removed by its author team. However, in order to demonstrate the generalization capability of our \name method across variable model sizes, we still include this model in the performance and efficiency tests here.

Table~\ref{tab:model_size_detail} provides the VideoMME scores for various model sizes across different video lengths and categories, offering a numerical breakdown of Figure~\ref{fig:scaling_model_size}.

\begin{table}[htb]
    \setlength\tabcolsep{3pt}
    \centering
    \begin{tabular}{l|cccc|r}
    \toprule
     & \multicolumn{4}{c|}{\textbf{Number of frames}} & \textbf{Max.}\\
    \textbf{Method} & \textbf{64} & \textbf{85} & \textbf{107} & \textbf{128}  & \textbf{Relative Drop}\\
    \midrule
    Original & 76.4 & 78.4 & 80.7 & 82.9 & - \\
    \midrule
    StreamingLLM & 23.3 & 25.8 & 27.6 & 27.6 & 70\% \\
    FastV & 58.2 & 63.6 & 65.8 & 69.3 & 24\% \\
    Ours & \textbf{75.3} & \textbf{78.2} & \textbf{80.0} & \textbf{83.6} & \textbf{1\%} \\
    \bottomrule
    \end{tabular}
    \caption{Numeric VideoNIAH retrieval accuracy of different methods across various frame counts.}
    \label{tab:frame_performance_comparison}
    \vspace{-15pt}
\end{table}

Table~\ref{tab:frame_performance_comparison} illustrates how retrieval accuracy scales with the number of input frames, complementing the insights from Figure~\ref{fig:input_length}. As shown, \name maintains consistent accuracy improvements across increasing frame numbers, matching the performance of the original model. In contrast, both StreamingLLM and FastV exhibit noticeable drops in accuracy.

\begin{table}[ht]
\centering
\begin{tabular}{l|cc|cc}
\toprule
\multirow{2}{*}{Setting} & \multicolumn{2}{c|}{Qwen2-VL-7B} & \multicolumn{2}{c}{InternVL-2.5-8B}\\
~ & Original & Ours & Original & Ours \\
\midrule
w/o sub & 55.9 & 58.4 & 63.1 & 62.3 \\
w/sub & 60.6 & 61.1 & 66.3 & 64.2 \\
\bottomrule
\end{tabular}
\caption{Performance comparison between original and Framefusion of Qwen2-VL-7B and InternVL-2.5-8B on VideoMME.}
\label{tab:model_performance} 
\vspace{-20pt}
\end{table}

We further test the performance of two extra models: Qwen2-VL-7B and InternVL-2.5-8B. As shown in Table~\ref{tab:model_performance}, our method performs well compared to original models on VideoMME.

\subsubsection{Retrieval Benchmark Details}
\label{sec:exp/retrieval_acc_detail}
\begin{figure*}[tb]
    \centering
    \includegraphics[width=\linewidth]{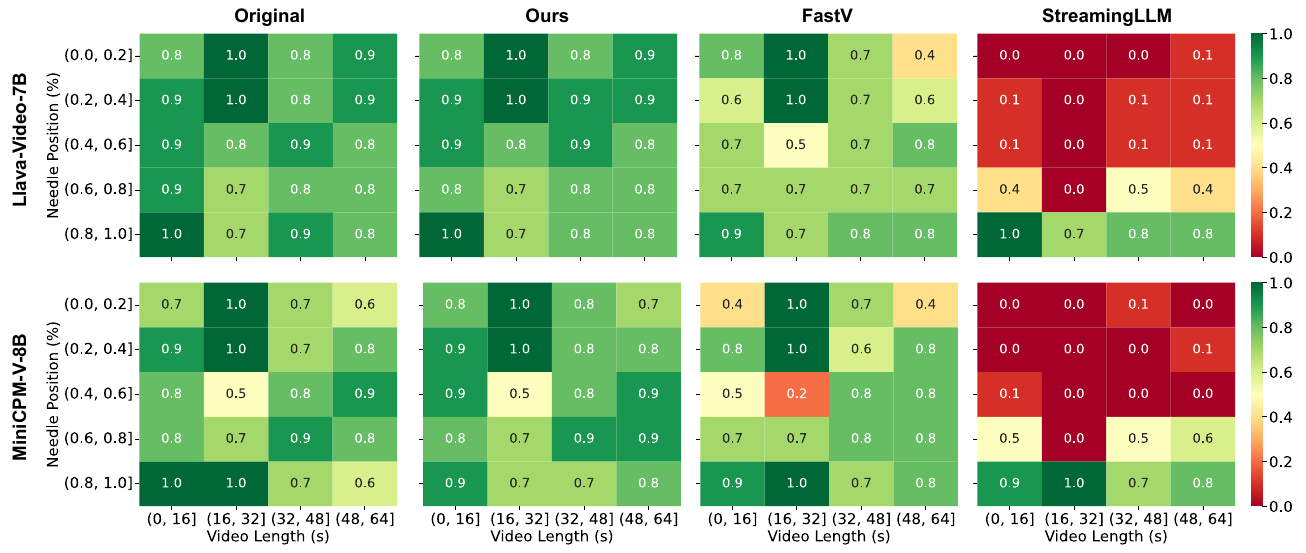}
    \caption{VideoNIAH retrieval accuracy of the Llava-Video-7B and MiniCPM-V-8B models using different token compression methods across varying video lengths and retrieval positions. All token compression methods employ 30\% relative token budget.
    }
    \label{fig:heatmap}
\end{figure*}

We further investigate the retrieval accuracy details with the VideoNIAH benchmark, as shown in Figure~\ref{fig:heatmap}.
\name demonstrates similar retrieval performance as the original dense model, with consistent performance across lengths and positions.
In contrast, StreamingLLM hardly retrieves the initial frames of the video. FastV does not show particular failure patterns but undergoes uniform performance degradation across grids.

\subsubsection{Performance on Image Benchmark}

We further investigate our method's performance on an image benchmark: MMMU-Pro-standard. As shown in Table~\ref{tab:nvila_results}, although our method is not designed for image inputs, it still demonstrates comparative performance.

\begin{table}[h]
\centering
\vspace{-3pt}
\begin{tabular}{l|c|c@{\hskip 6pt}c}
\toprule
\textbf{Model} & \textbf{Original} & \textbf{Fastv} & \textbf{Ours} \\
\midrule
NVILA-2B  & 23.6 & \textbf{23.8} & 23.1 \\
NVILA-8B  & 30.3 & 28.7 & \textbf{29.0} \\
NVILA-15B & 36.1 & 30.6 & \textbf{32.8} \\
\bottomrule
\end{tabular}
\caption{The MMMU-Pro-standard performance across NVILA-2B, 8B, and 15B for different methods.}
\label{tab:nvila_results}
\end{table}

\subsection{Efficiency}
\label{sec:appendix/efficiency}

\subsubsection{Efficiency Across Different Model Sizes}
\label{sec:speedup_model_size}

\begin{figure}[tb]
    \centering
    \includegraphics[width=\linewidth]{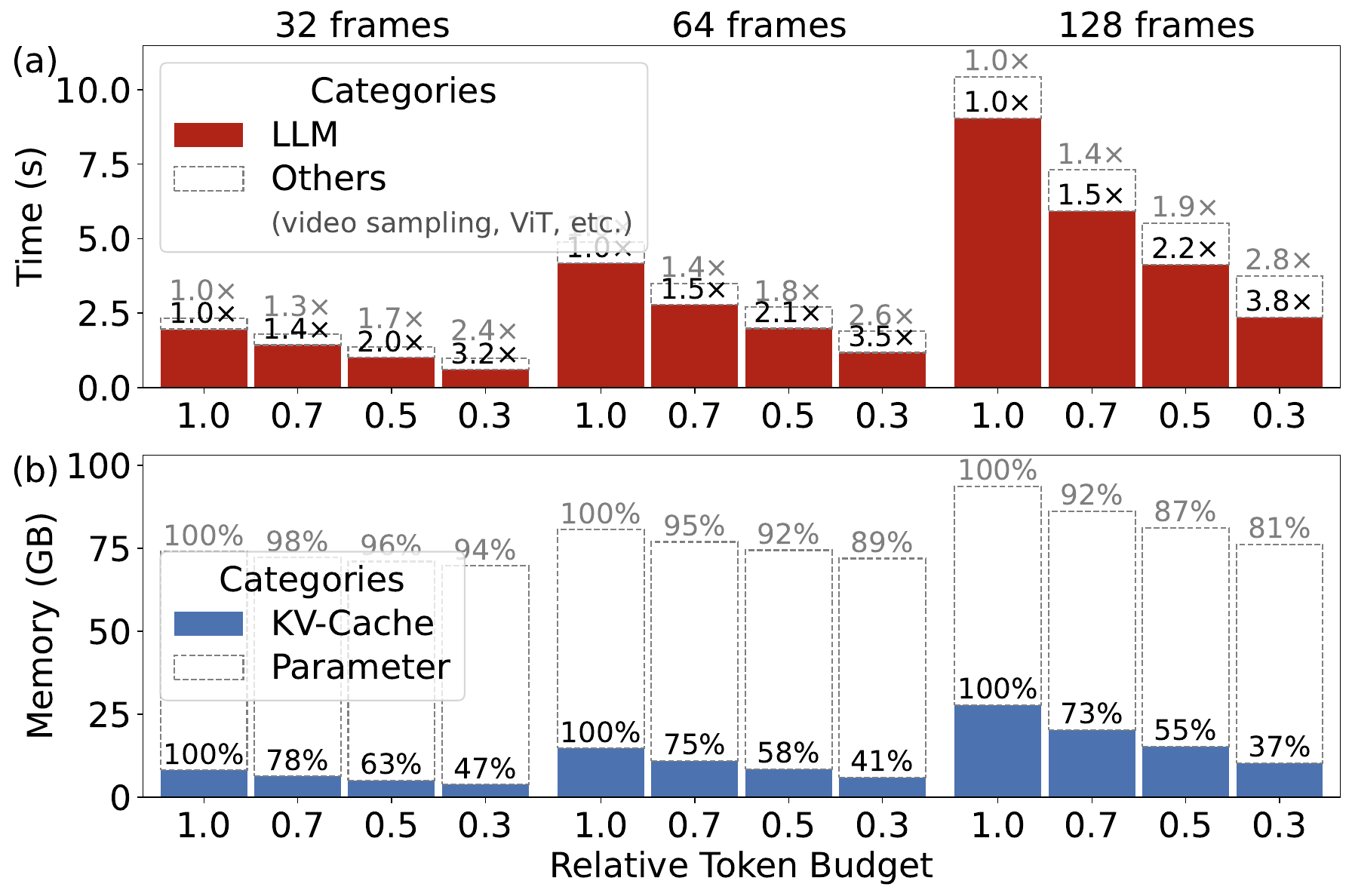}
    \caption{Runtime and memory breakdown of Llava-Video-32B on two A100-80GB GPUs using \name. A relative token budget of 1.0 represents the original dense model. Numbers on bars show (a) LLM and end-to-end speedups and (b) LLM's KV-Cache and total relative memory.}
    \label{fig:efficiency_32B}
\end{figure}

\begin{figure}[tb]
    \centering
    \includegraphics[width=\linewidth]{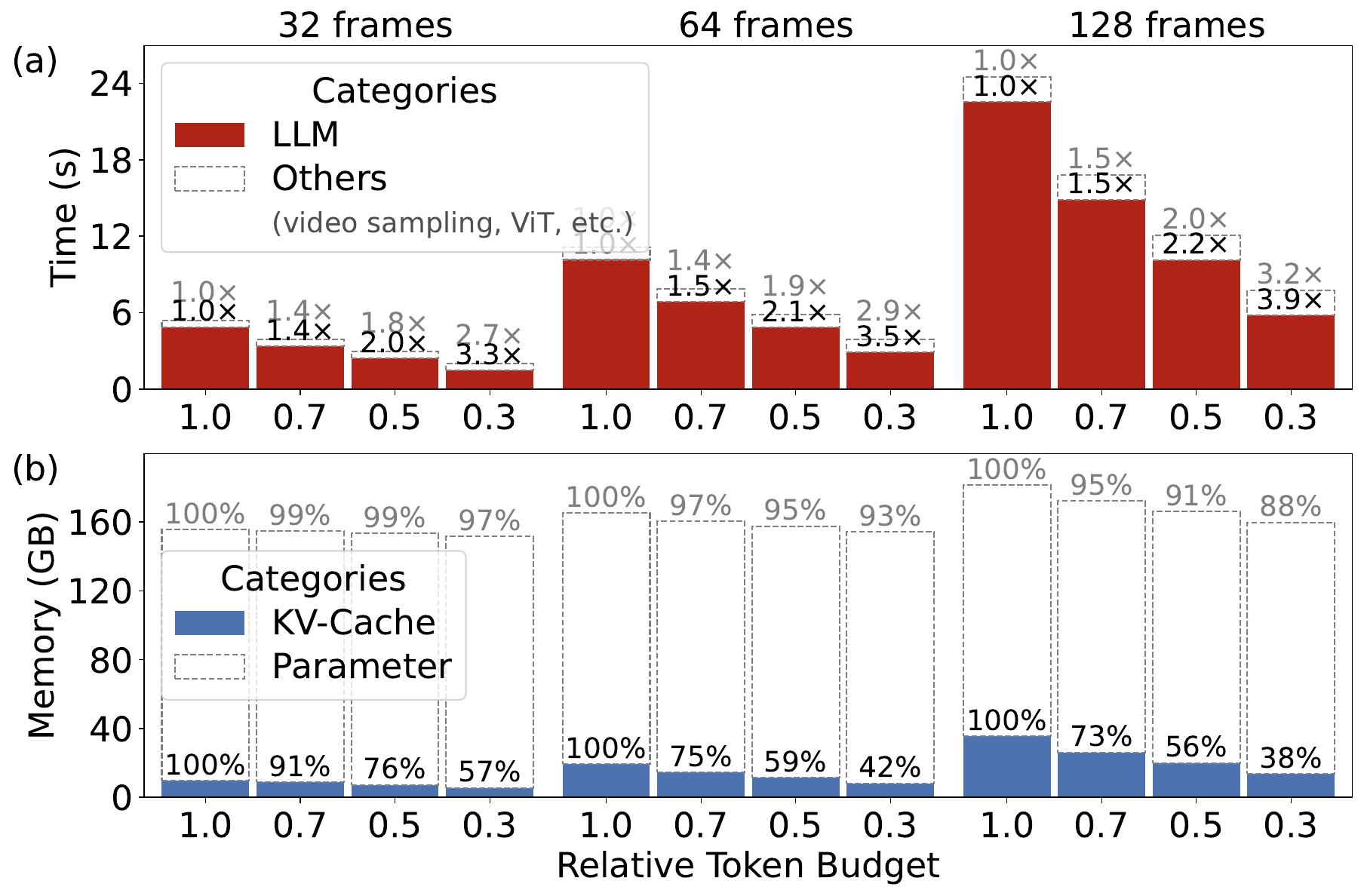}
    \caption{Runtime and memory breakdown of Llava-Video-72B on four A100-80GB GPUs using \name. A relative token budget of 1.0 represents the original dense model. Numbers on bars show (a) LLM and end-to-end speedups and (b) LLM's KV-Cache and total relative memory.}
    \label{fig:efficiency_72B}
\end{figure}

We evaluate the scalability of \name’s efficiency across different model sizes, as shown in Figure~\ref{fig:efficiency_32B} and \ref{fig:efficiency_72B}. To accommodate the increased KV-Cache and memory overhead, we distribute models across multiple GPUs. With larger models, \name achieves greater end-to-end speedups, delivering $2.8\times$ for Llava-Video-32B on two GPUs and $3.2\times$ for Llava-Video-72B on four GPUs at a 30\% token budget.
Besides, \name reduces memory consumption for KV-Cache to 37\% for Llava-Video-32B and 51\% for Llava-Video-72B with a 30\% token budget.

\begin{figure}
    \centering
    \includegraphics[width=\linewidth]{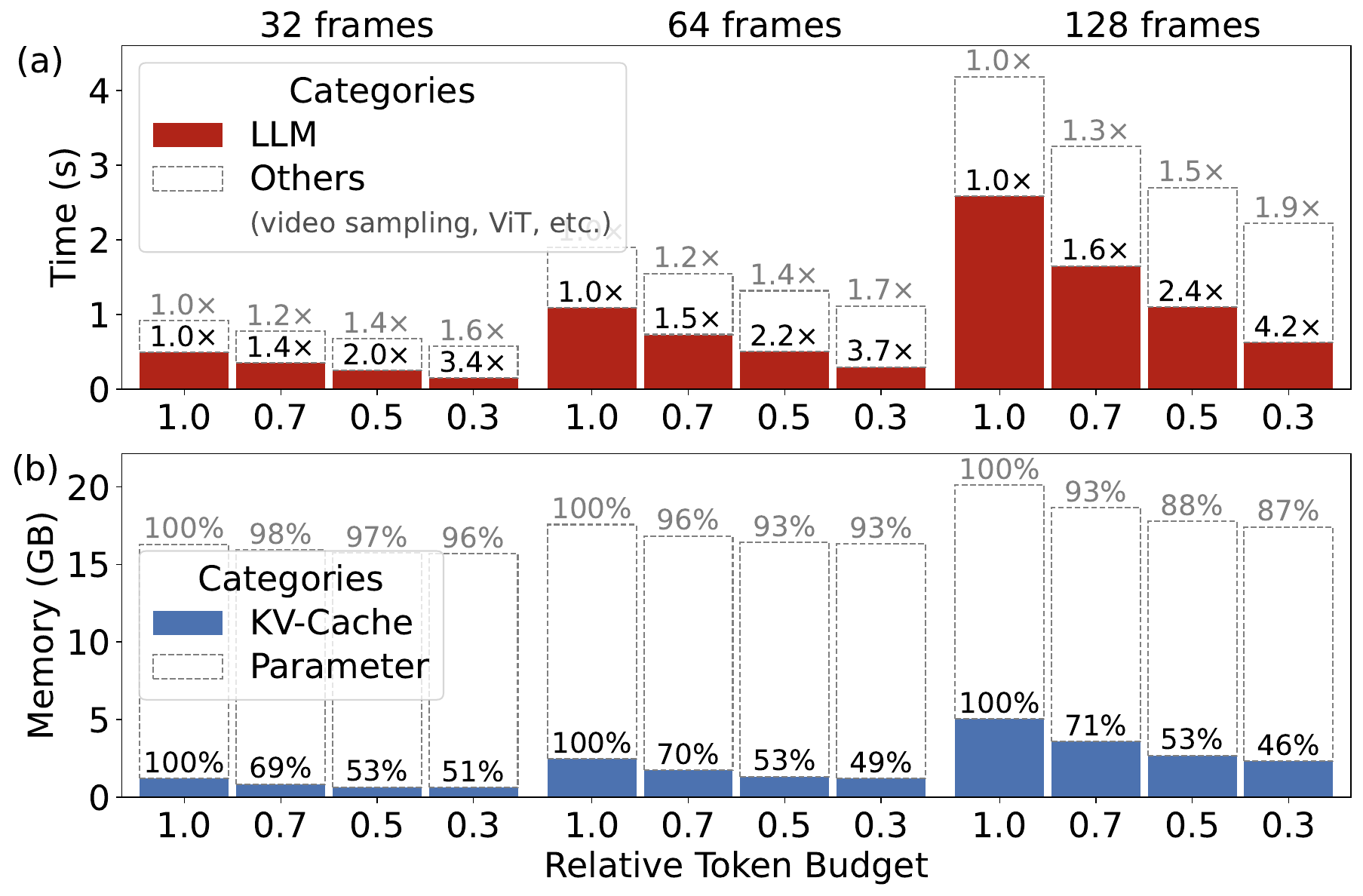}
    \caption{Runtime and memory breakdown of Llava-Video-7B on a single A100-80GB GPU using \name. A relative token budget of 1.0 represents the original dense model. Numbers on bars show (a) LLM and end-to-end speedups and (b) LLM's KV-Cache and total relative memory.}
    \label{fig:exp/speedup}
\end{figure}

\subsubsection{Token Reduction Details}


\begin{figure}
    \centering
    \includegraphics[width=\linewidth]{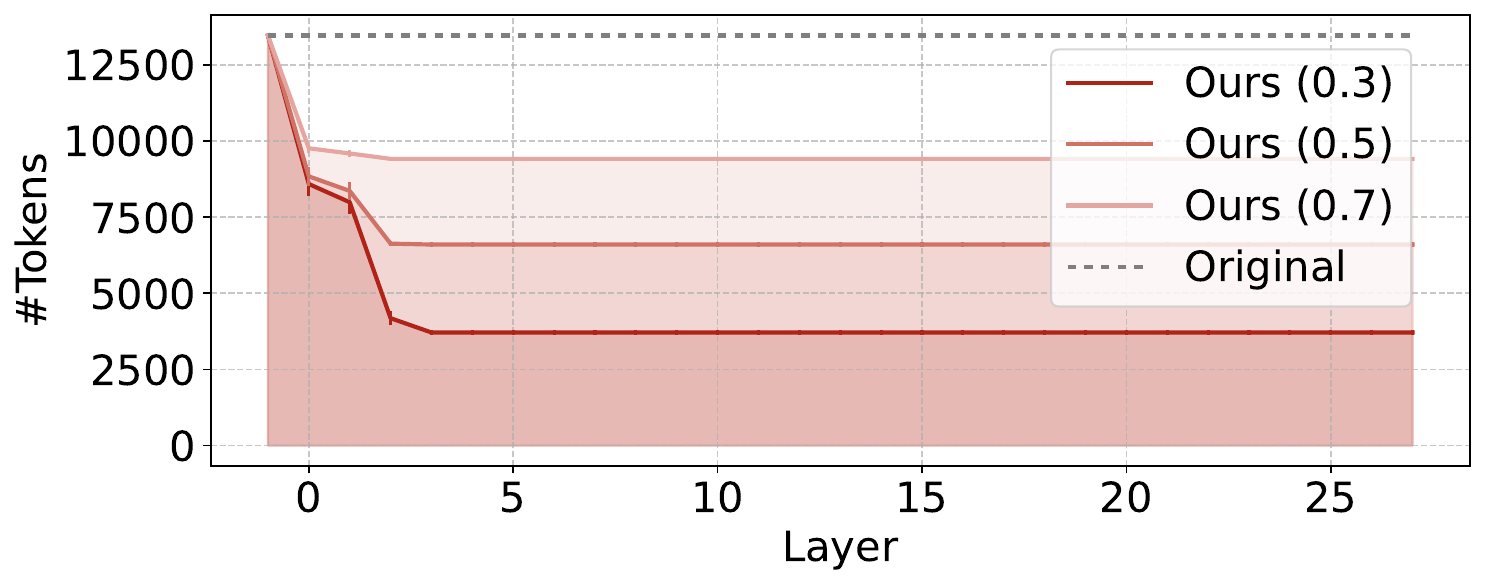}
    \caption{
        Average number of tokens per layer in the Llava-Video-7B model with \name at different relative token budgets. Error bars represent variance across data items.
    }
    \label{fig:token_num_per_layer}
\end{figure}

\name reduces computational cost through both token merging and pruning. Using 128 samples from the VideoMME dataset with the Llava-Video-7B model, we calculate the token count per layer. As shown in Figure~\ref{fig:token_num_per_layer}, \name progressively reduces tokens per layer, achieving the desired relative token budget (represented by the area under the line).

\subsection{Ablation Study}
\label{sec:appendix/ablation}

\subsubsection{Similarity Computation Strategy}

\begin{figure}
    \centering
    \includegraphics[width=0.8\linewidth]{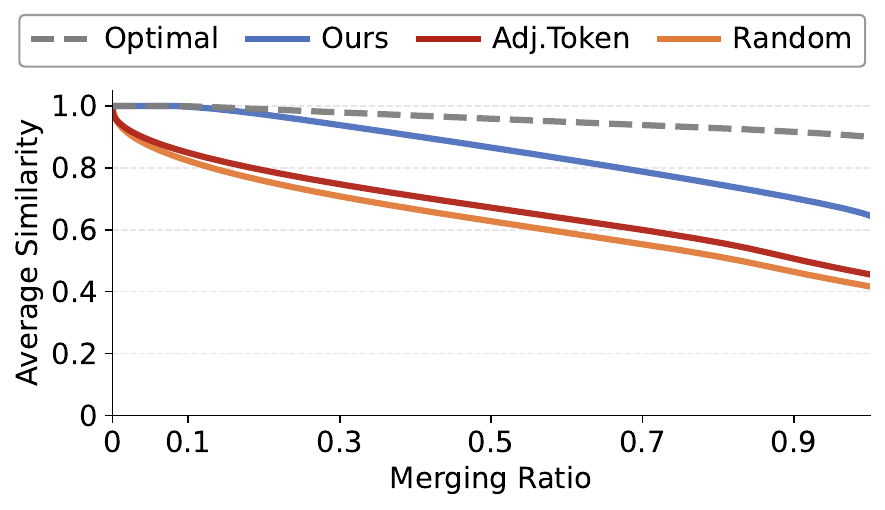}
    \caption{The average token similarity of the merged tokens for the first layer of Llava-Video-7B model across various merging rates.}
    \label{fig:similarity_hit}
\end{figure}

We empirically study whether our approach successfully finds the most similar token pairs. 
All three $O(N)$ complexity strategies are compared against the posterior optimal upper bound, which merges the most similar tokens using the full $N \times N$ similarity computation.
As shown in Figure~\ref{fig:similarity_hit}, given different merging rate, the token pairs found by our method constantly shows the highest average similarity. We successfully reach 90\% average similarity with only $1/10^4$ computing overhead. Further ablations are detailed in Section~\ref{sec:ablation}

\subsubsection{Distance Metrics}
\label{sec:distance_metrices}

\name adopts cosine similarity as the distance metric between tokens.
To evaluate the impact of different distance metrics, we replace cosine similarity with the inner product, Minkowski-2, and Minkowski-1 distance. We test the performance of \name at a 30\% token budget.
As shown in Table~\ref{tab:distance_strategy}, the average accuracy using cosine similarity is 2.8\%, 1.4\%, and 1.4\% higher than the baseline metrics, respectively.

\begin{table}[tb]
    \centering
    \setlength\tabcolsep{1pt} 
    \begin{tabular}{l|ccc|c}
    \toprule
    \textbf{Choice} & \textbf{VideoNIAH} & \textbf{VideoMME} & \textbf{NExt-QA} & \textbf{Avg.}\\
    \toprule
    inner product  & 71.3 & 58.9 & 55.0 & 61.7 \\
    minkowski-2 & 71.3& 60.9 & \textbf{57.1} & 63.1 \\
    minkowski-1 & 71.3  & 61.0 & 57.0 & 63.1 \\
    cosine similarity & \textbf{75.1} & \textbf{61.4} & 56.9 & \textbf{64.5} \\
    \bottomrule
    \end{tabular}
    \caption{Performance of different distance calculation strategies with the same relative token budget of 30\% on VideoNIAH, VideoMME, and NExt-QA.}
    \label{tab:distance_strategy}
\end{table}

\subsubsection{Choice of Similarity Threshold}
\label{sec:ablation_threshold}
We conduct ablation studies on the sensitivity of the similarity ($S_{\text{threshold}}$) and merging-pruning transition ($N_{\text{threshold}}$) thresholds on NVILA-8B.
As shown in Table \ref{tab:ablation_threshold}, our method shows robust performance to threshold variations.

\begin{table}[h]
\centering
\vspace{-8pt}
\setlength\tabcolsep{2pt} 
\resizebox{\linewidth}{!}{
\begin{tabular}{ll|ccc}
\toprule
\textbf{Threshold} & \textbf{Value} & \textbf{VideoNIAH} & \textbf{VideoMME} & \textbf{NeXT-QA-mc} \\
\midrule
\multirow{4}{*}{$S_{\text{threshold}}$} 
    & 0.6 & 73.6 & 56.9 & 56.5 \\
    & 0.7 (default) & 73.3 & 57.4 & 56.3 \\
    & 0.8 & 72.9 & 57.6 & 56.5 \\
    & 0.9 & 72.2 & 57.7 & 56.3 \\
\midrule
\multirow{3}{*}{$N_{\text{threshold}}$} 
    & 0.1 (default) & 73.3 & 57.4 & 56.3 \\
    & 0.2 & 74.0 & 57.6 & 56.5 \\
    & 0.3 & 74.2 & 57.5 & 56.5 \\
\bottomrule
\end{tabular}
}
\caption{Performance of different similarity and merging-pruning transition thresholds on VideoNIAH, VideoMME, and NExt-QA.}
\label{tab:ablation_threshold}
\end{table}

\subsubsection{Effect of Positional Embedding}
We investigate the impact of positional embeddings on token similarity.
Specifically, we compare models with and without positional embedding at the first layer and analyze the resulting changes in the similarity of the input hidden states to the second layer.
The results show that the L1-norm of the similarity matrix changes by an absolute amount of $0.0087\pm0.0010$, corresponding to a relative change of $2.73\%\pm0.66\%$. It shows that the token contents, rather than the positional embeddings, dominate token similarity.

\section{Asymptotic Complexity Analysis}
\label{sec:appendix/asymptotic_complexity}

We estimate the computing cost of \name following the approach of FastV~\citep{chen2024fastv}. 
Given a model with $L$ layers and a specified relative token budget $C$, \name operates in the merging stage from layer 0 to layer $K - 1$, then transitions to the pruning stage at layer $K$. 
Let $N_l$ denote the number of tokens in layer $l$ before token reduction at this layer. Note that $N_{l+1}$ represents the number of tokens of layer $l$ after token reduction, and we let $N_{-1}$ equal the original input token length $N$.
\name reduces $N_l$ with merging and pruning at the initial $K+1$ layers. After the token reduction, the remaining tokens for the successive layers are calculated as follows:
\begin{equation}
    N_{l} = \frac{L \times C \times N - (N_0 + \ldots + N_{K})}{L - K -1}, l \in [K+1, L)
    \label{eq:token_num}
\end{equation}
    
The model inference computation FLOPs $F(N_{l}, N_{l+1})$ of layer $l$ is calculated as follows:
\begin{equation}
    F(N_{l}, N_{l+1}) = 4N_l D^2 + 2N_l^2 D + 3N_{l+1} D M
    \label{eq:compute_cost}
\end{equation}
where $D$ denotes the hidden state size, and $M$ denotes the intermediate FFN size.
The additional computation $F'(N_{l})$ introduced by \name during similarity computation is:
\begin{equation}
    F'(N_l) = 3N_{l} D
    \label{eq:merge_cost}
\end{equation}
Note that the additional computation $F'$ introduced by \name shows negligible asymptotic complexity with respect to input length and model size, compared with the $O(N^2D)$ and $O(ND^2)$ complexities of the original model. 

\section{Additional Observation Details}
\label{sec:appendix/observation}

\subsection{Similarity Distribution Details}
\label{sec:appendix/similarity_distribution}

\begin{figure}
    \centering
    \includegraphics[width=0.93\linewidth]{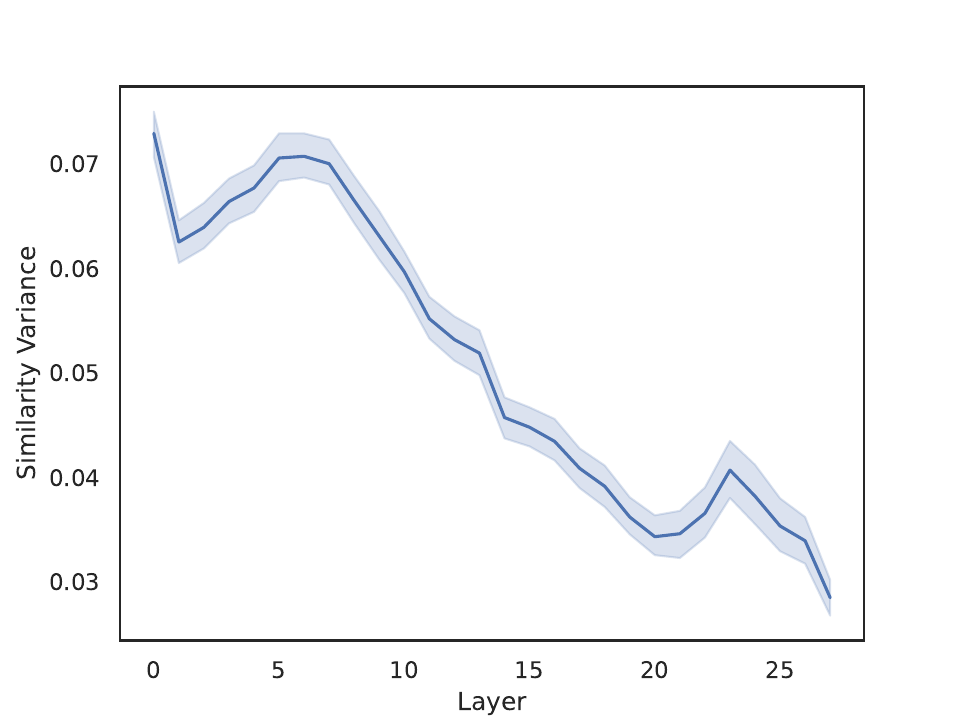}
    \caption{Average token similarity variance per LLM layer in the Llava-Video-7B model, tested on 128 samples from the VideoMME dataset. Shading represents the variance across data items.}
    \label{fig:similarity_variance}
\end{figure}

We take 128 videos from the VideoMME dataset and calculate the variance in token similarity across different layers. 
As shown in Figure~\ref{fig:similarity_variance}, the similarity variance decreases in the deeper layers of the model, validating Observation~\ref{obs:shift}.
No significant outliers are observed in token similarity, in contrast to the common outliers seen with respect to the magnitude of hidden features~\citep{zhang2025sageattention, zhang2024sageattention2}.

\subsection{Observations on Additional Models}
\label{sec:observe_minicpm}

\begin{figure}
    \centering
    \includegraphics[width=0.8\linewidth]{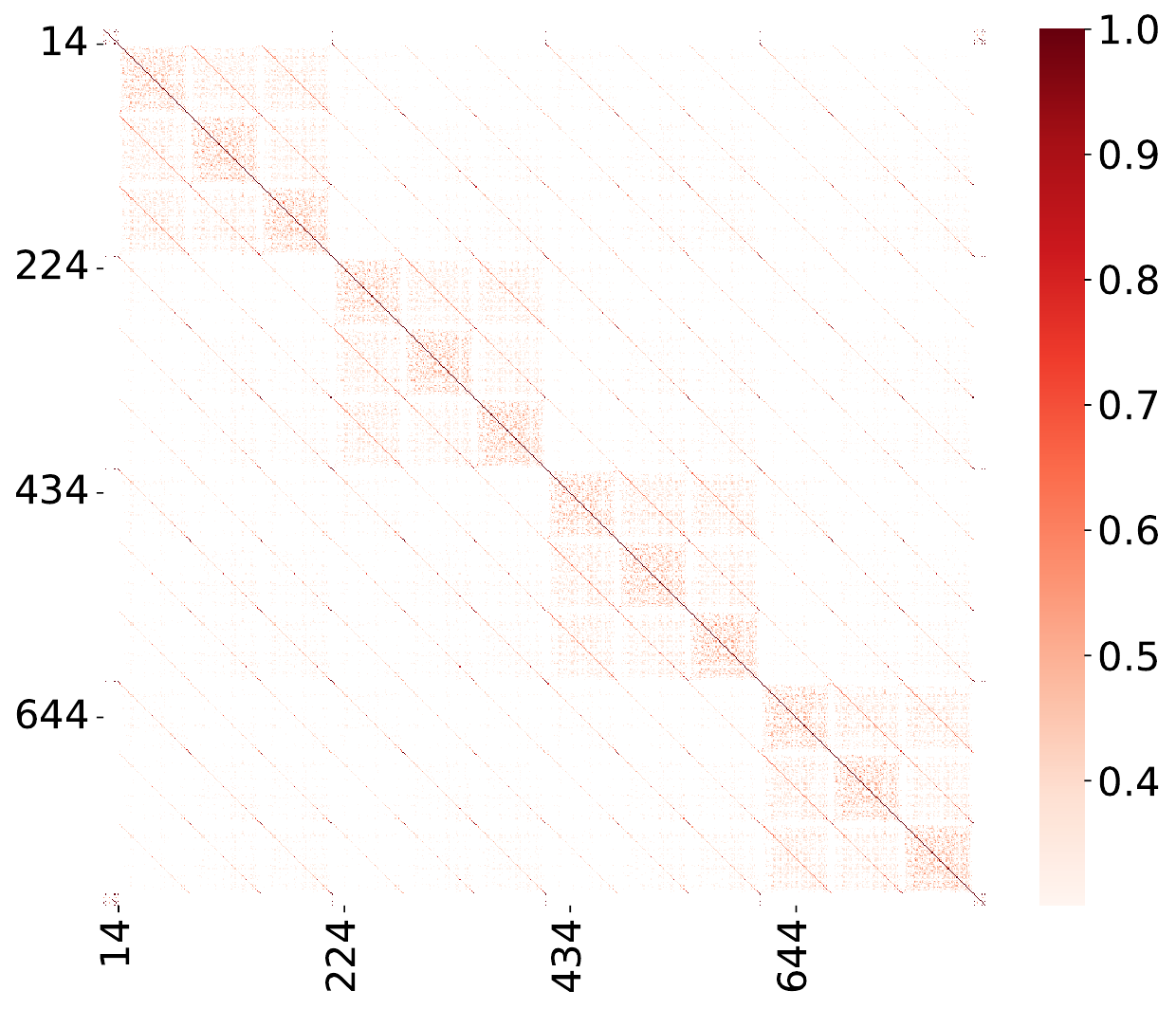}
    \caption{Token similarities between all input tokens at the first LVLM layer in MiniCPM-V-8B.}
    \label{fig:token_similarity_minicpmv}
\end{figure}

\begin{figure}
\centering
\includegraphics[width=\linewidth]{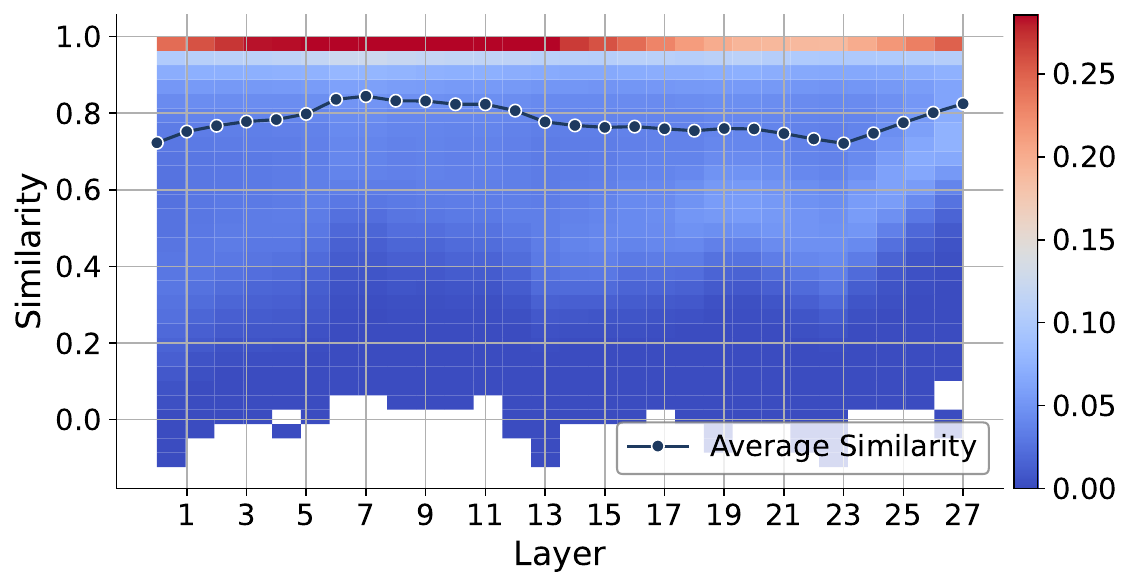}
\caption{
Heatmap of token similarity across different model layers for the MiniCPM-V-8B model. 
Each cell represents the similarity at a specific layer, with color intensity denoting distribution frequency.
The line overlay shows the average token similarity across layers.
}
\label{fig:similarity_distribution_minicpmv}
\end{figure}

\begin{figure}
    \centering
    \includegraphics[width=\linewidth]{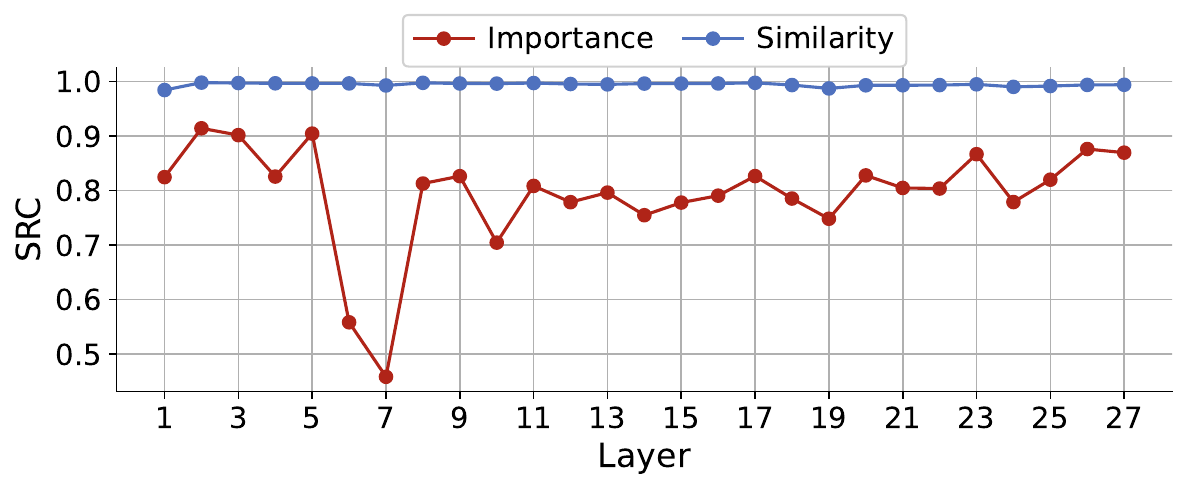}
    \caption{Spearman Rank Correlation (SRC) between adjacent layers for the MiniCPM-V-8B model.}
    \label{fig:spearman_rank_minicpmv}
\end{figure}

\begin{figure}[htbp]
    \centering
    \begin{subfigure}[b]{\linewidth}
        \centering
        \includegraphics[width=\linewidth]{fig/retention_ratio_label.pdf}
        \vspace{-13pt}
    \end{subfigure}

    \begin{subfigure}[b]{\linewidth}
        \centering
        \includegraphics[width=\linewidth]{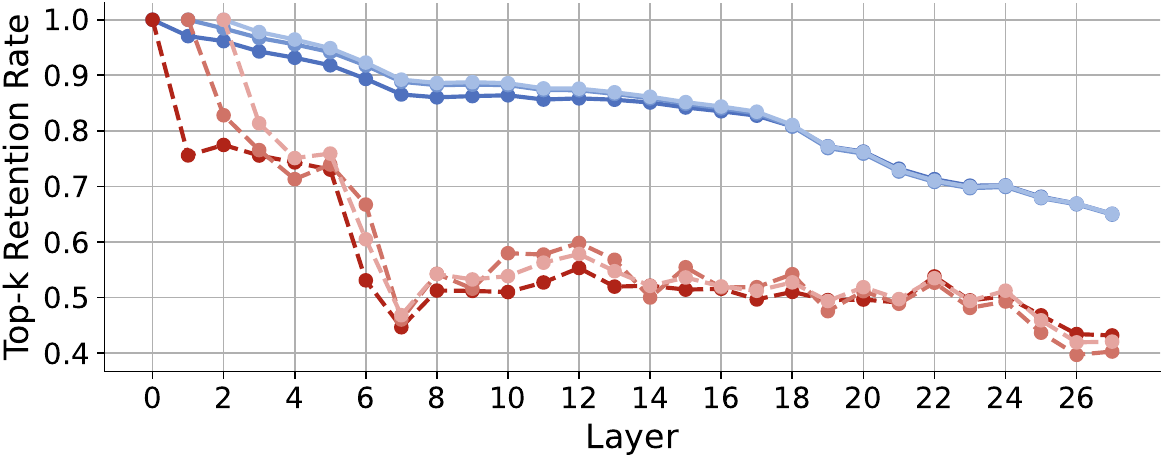}
        \vspace{-8pt}
    \end{subfigure}

    \caption{The Top-30\% retention rate across model layers for the MiniCPM-V-8B model, using different retention metrics and reference layers.}
    \label{fig:retention_rate_plot_minicpmv}
\end{figure}

In addition to the analysis of the Llava-Video model in Section~\ref{sec:analysis}, we conduct a similar study on the MiniCPM architecture. Results are presented in Figures~\ref{fig:token_similarity_minicpmv}, \ref{fig:similarity_distribution_minicpmv}, \ref{fig:spearman_rank_minicpmv}, and \ref{fig:retention_rate_plot_minicpmv}.

Overall, the conclusions align with those of the Llava-Video model, with a few notable differences:
Firstly, as shown in Figure~\ref{fig:token_similarity_minicpmv}, MiniCPM, which incorporates Q-Former~\citep{minicpm, li2023blip2}, exhibits additional high similarity among visual tokens within the same frame. However, the prominent 210th sub-diagonal persists, supporting our token similarity calculation strategy.
Secondly, as shown in Figure~\ref{fig:similarity_distribution_minicpmv}, high similarity decreases less steeply in deeper layers for MiniCPM compared to Llava-Video. Despite this, the superior efficiency of cascaded merging at shallower layers ensures that Design Choice~\ref{dsn:merge_then_prune} remains valid.

\subsection{Video Pruning Visualization}

\begin{figure*}[htbp]
\centering
\foreach \n in {0,...,63} {
    \begin{subfigure}{0.12\textwidth}
        \centering
        \includegraphics[width=\linewidth]{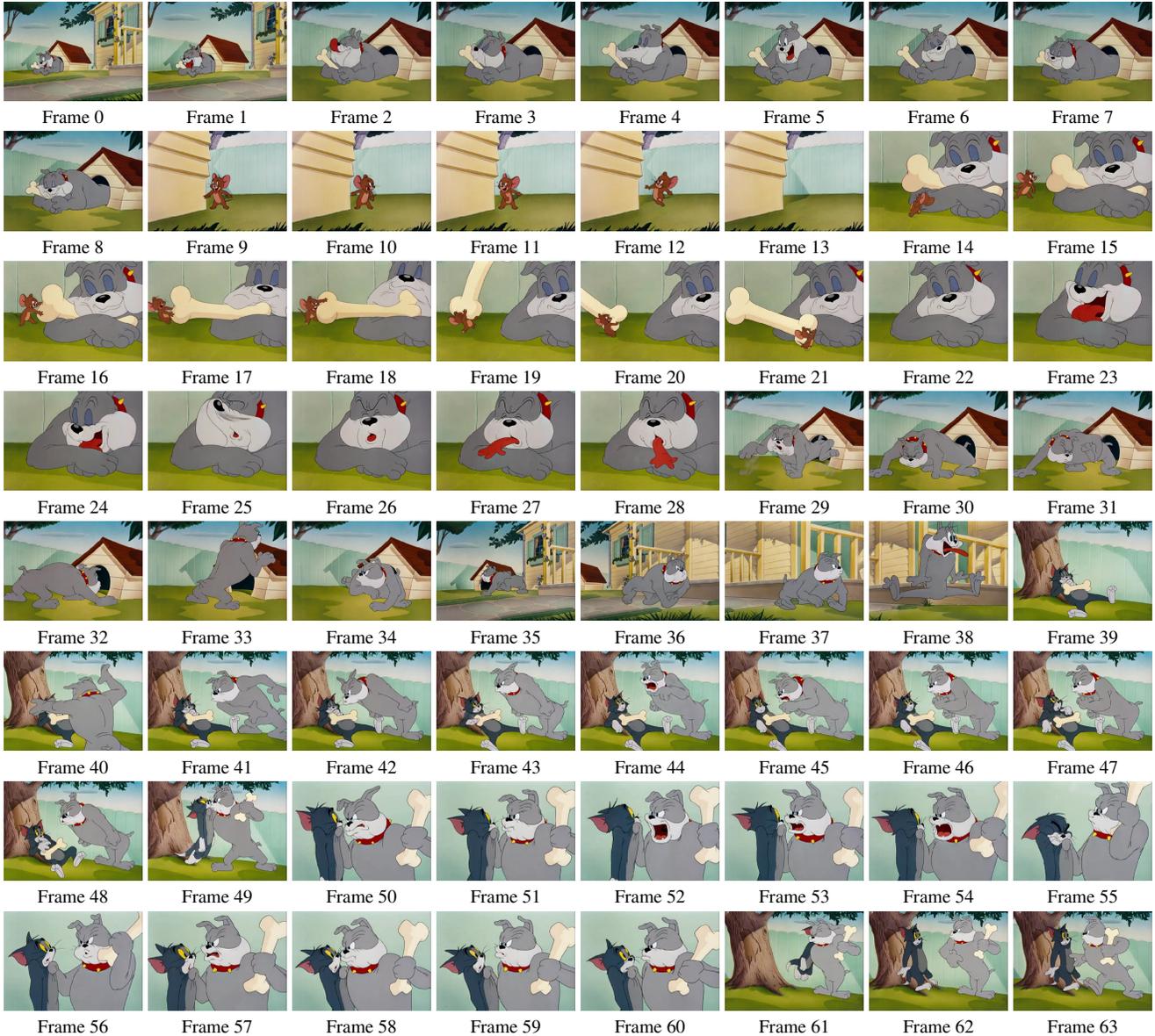} 
        \captionsetup{skip=3pt} 
        \caption*{Frame \n}
    \end{subfigure}%
    \ifnum\n=7 \par\fi 
    \ifnum\n=15 \par\fi
    \ifnum\n=23 \par\fi
    \ifnum\n=31 \par\fi
    \ifnum\n=39 \par\fi
    \ifnum\n=47 \par\fi
    \ifnum\n=55 \par\fi
    \ifnum\n=63 \par\fi
}
\caption{An example input video with 1 fps frame rate.}
\label{fig:original_example}
\end{figure*}

\begin{figure*}[htbp]
\centering
\foreach \n in {0,...,63} {
    \begin{subfigure}{0.12\textwidth}
        \centering
        \includegraphics[width=\linewidth]{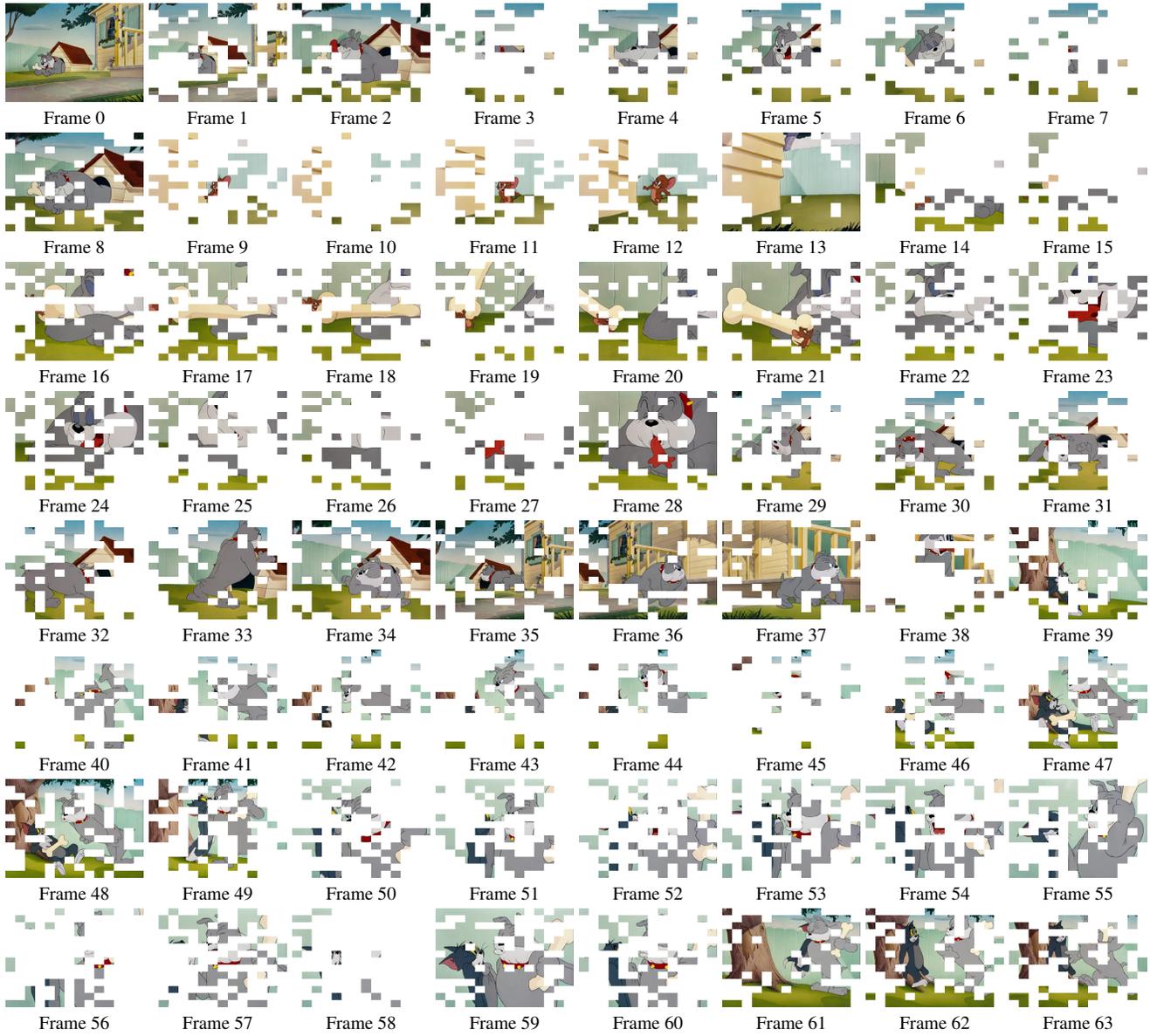} 
        \captionsetup{skip=3pt} 
        \caption*{Frame \n}
    \end{subfigure}%
    \ifnum\n=7 \par\fi 
    \ifnum\n=15 \par\fi
    \ifnum\n=23 \par\fi
    \ifnum\n=31 \par\fi
    \ifnum\n=39 \par\fi
    \ifnum\n=47 \par\fi
    \ifnum\n=55 \par\fi
    \ifnum\n=63 \par\fi
}
\caption{The example of the video after token merging. Merged tokens are visualized with the blank blocks.}
\label{fig:example_merge_blank}
\end{figure*}

\begin{figure*}[htbp]
\centering
\foreach \n in {0,...,63} {
    \begin{subfigure}{0.12\textwidth}
        \centering
        \includegraphics[width=\linewidth]{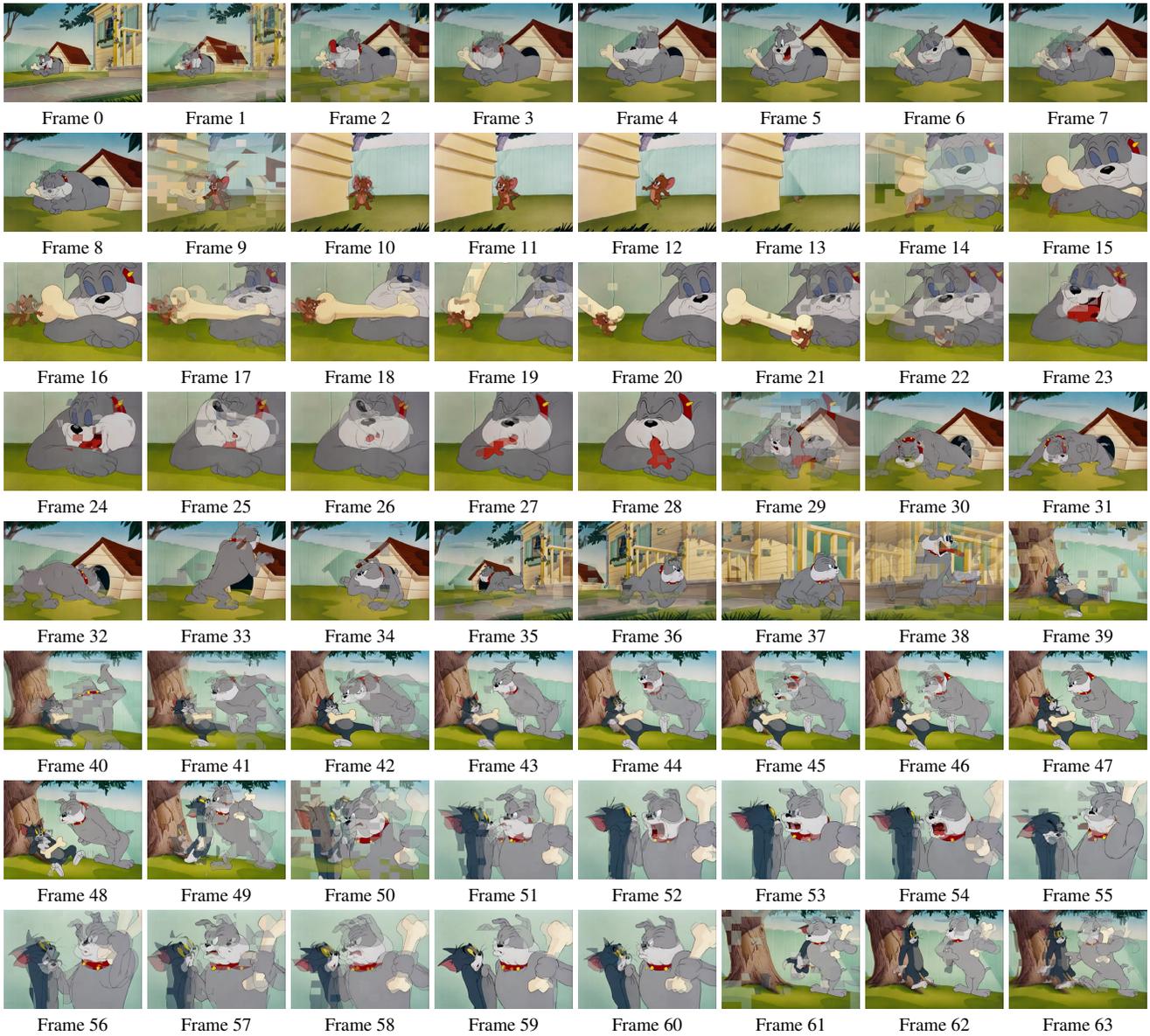} 
        \captionsetup{skip=3pt} 
        \caption*{Frame \n}
    \end{subfigure}%
    \ifnum\n=7 \par\fi 
    \ifnum\n=15 \par\fi
    \ifnum\n=23 \par\fi
    \ifnum\n=31 \par\fi
    \ifnum\n=39 \par\fi
    \ifnum\n=47 \par\fi
    \ifnum\n=55 \par\fi
    \ifnum\n=63 \par\fi
}
\caption{The example of the video after token merging. Merged tokens are visualized with the average image patches.}
\label{fig:example_merge_average}
\end{figure*}

We select a video example to visualize the effect of our token merging strategy. 
Figure~\ref{fig:original_example} shows the frames of the original video sampled at a frame rate of 1 fps. 
In Figure~\ref{fig:example_merge_blank}, we present the video input to the model after token merging in Layer 0, where blank patches indicate tokens that have been merged. 
Furthermore, we replace the blank regions with the average of the merged patches, and the resulting visualization is shown in Figure~\ref{fig:example_merge_average}.
As shown in the examples, \name token merging strategy successfully merges similar visual tokens, reducing the computational costs, while maintaining high validity of the video.

\subsection{Importance-Similarity Joint-Distribution}
\begin{figure*}[htbp]
    \centering
    \begin{subfigure}{0.3\textwidth}
        \centering
        \includegraphics[width=\linewidth]{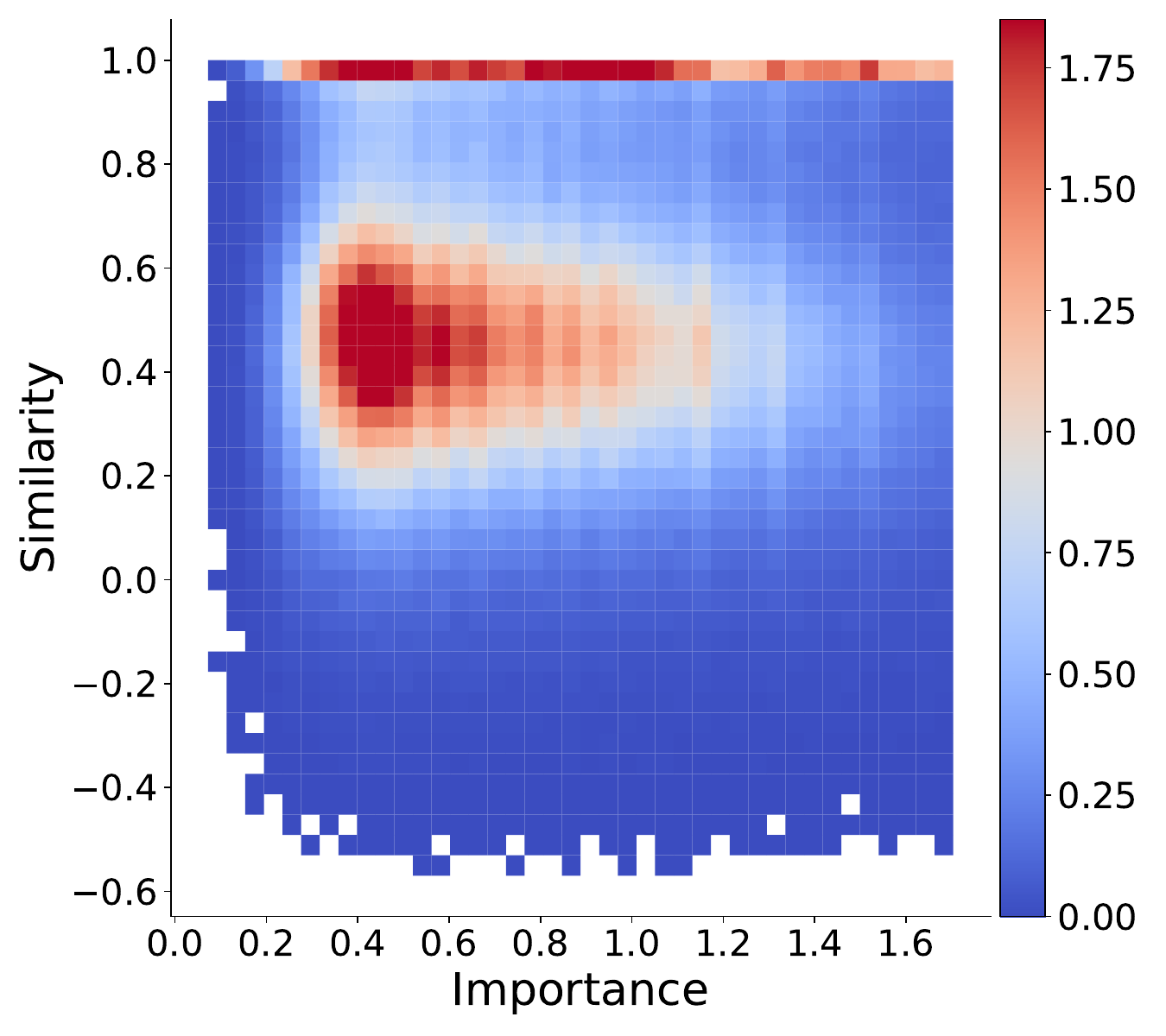}
        \caption{Layer 0}
    \end{subfigure}
    \begin{subfigure}{0.3\textwidth}
        \centering
        \includegraphics[width=\linewidth]{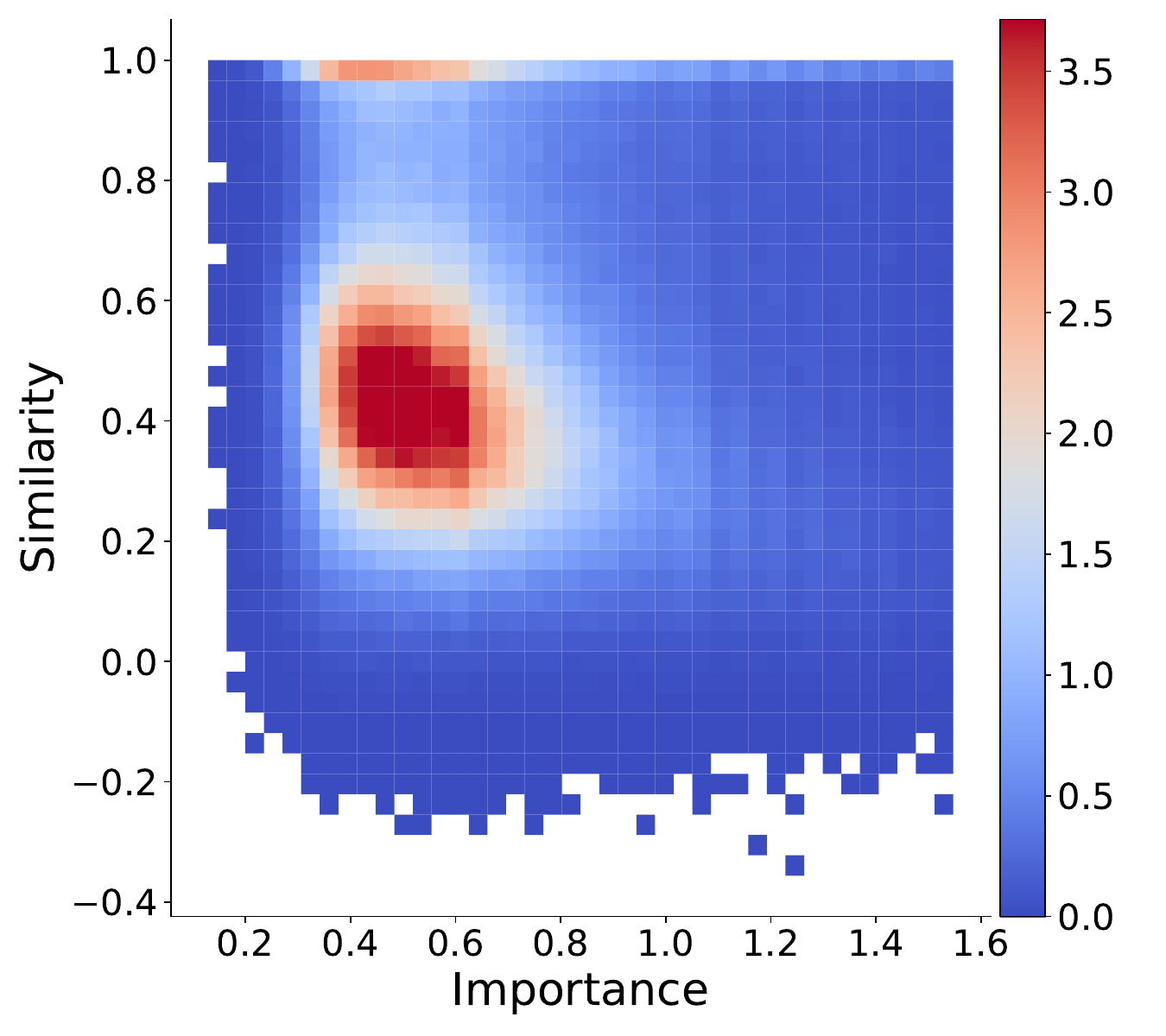}
        \caption{Layer 1}
    \end{subfigure}
    \begin{subfigure}{0.3\textwidth}
        \centering
        \includegraphics[width=\linewidth]{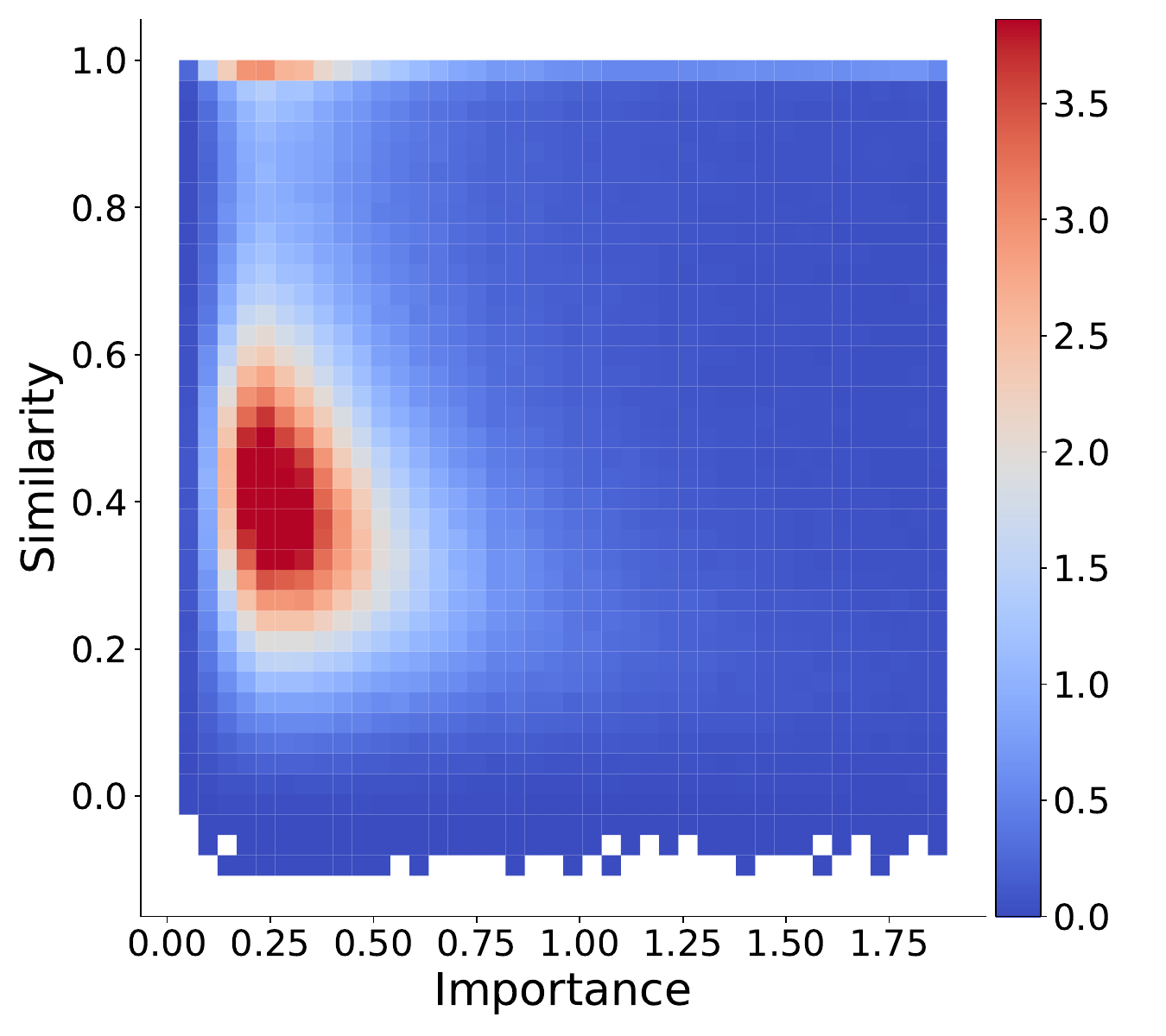}
        \caption{Layer 2}
    \end{subfigure}

    \begin{subfigure}{0.3\textwidth}
        \centering
        \includegraphics[width=\linewidth]{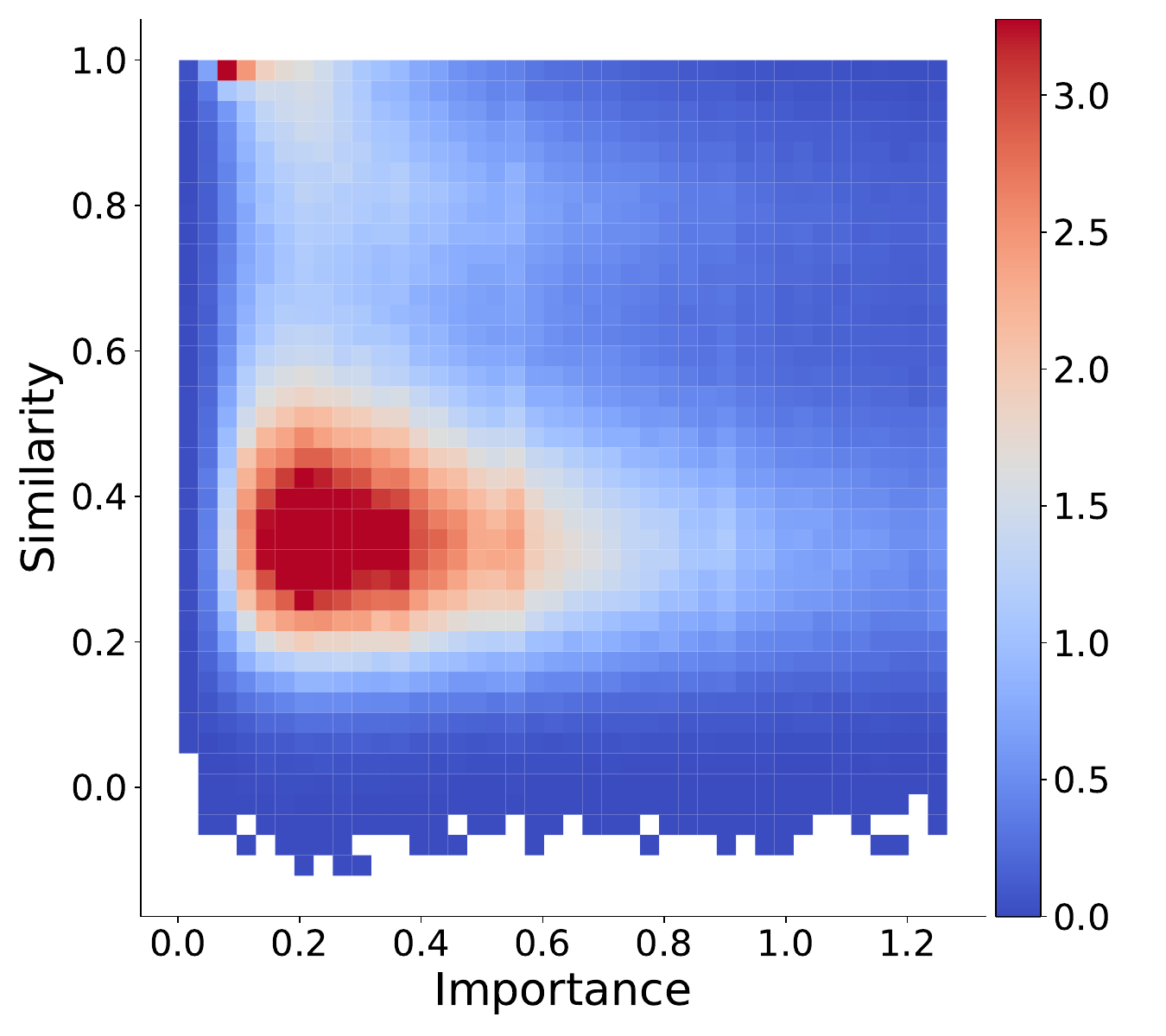}
        \caption{Layer 14}
    \end{subfigure}
    \begin{subfigure}{0.3\textwidth}
        \centering
        \includegraphics[width=\linewidth]{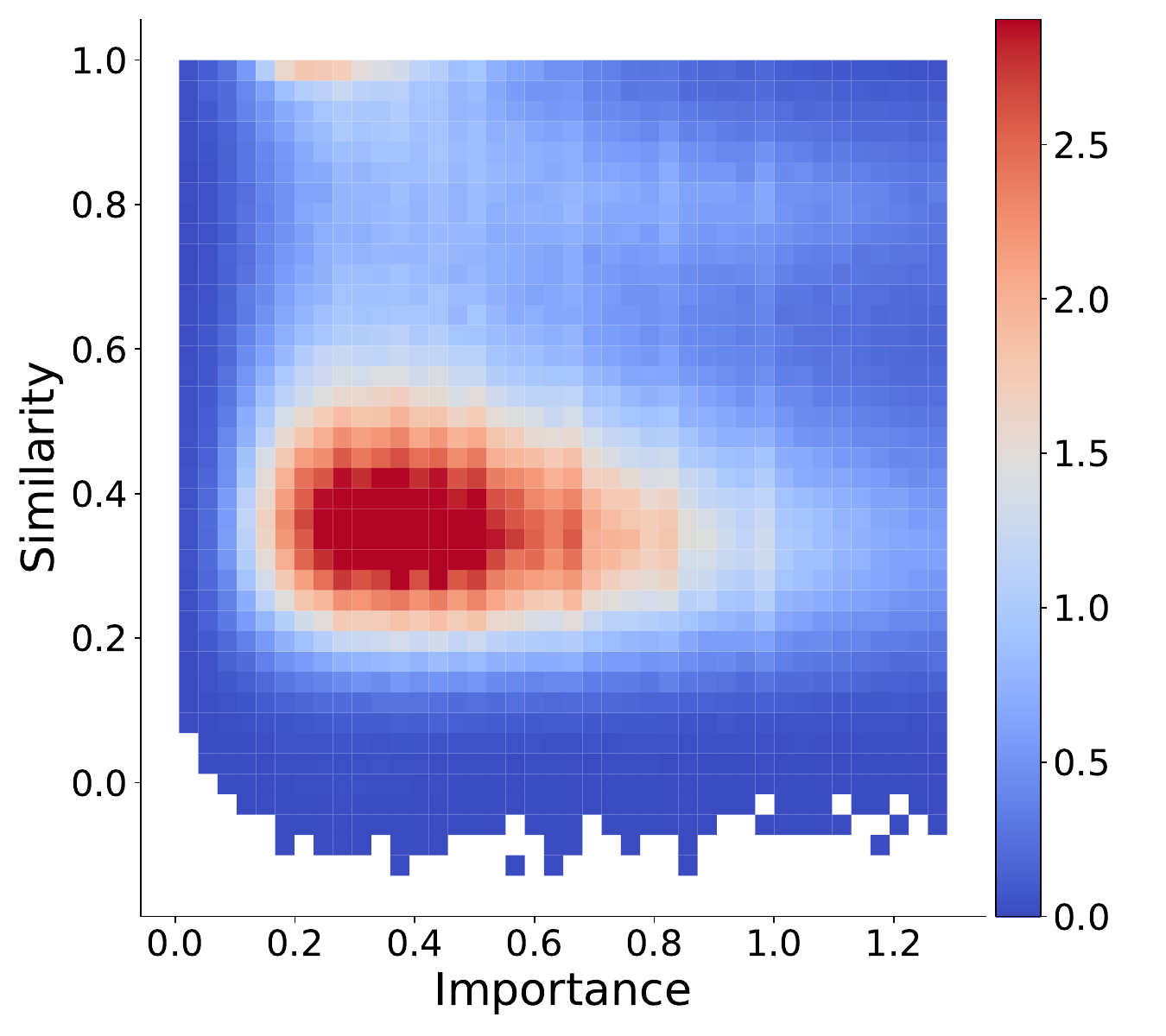}
        \caption{Layer 15}
    \end{subfigure}
    \begin{subfigure}{0.3\textwidth}
        \centering
        \includegraphics[width=\linewidth]{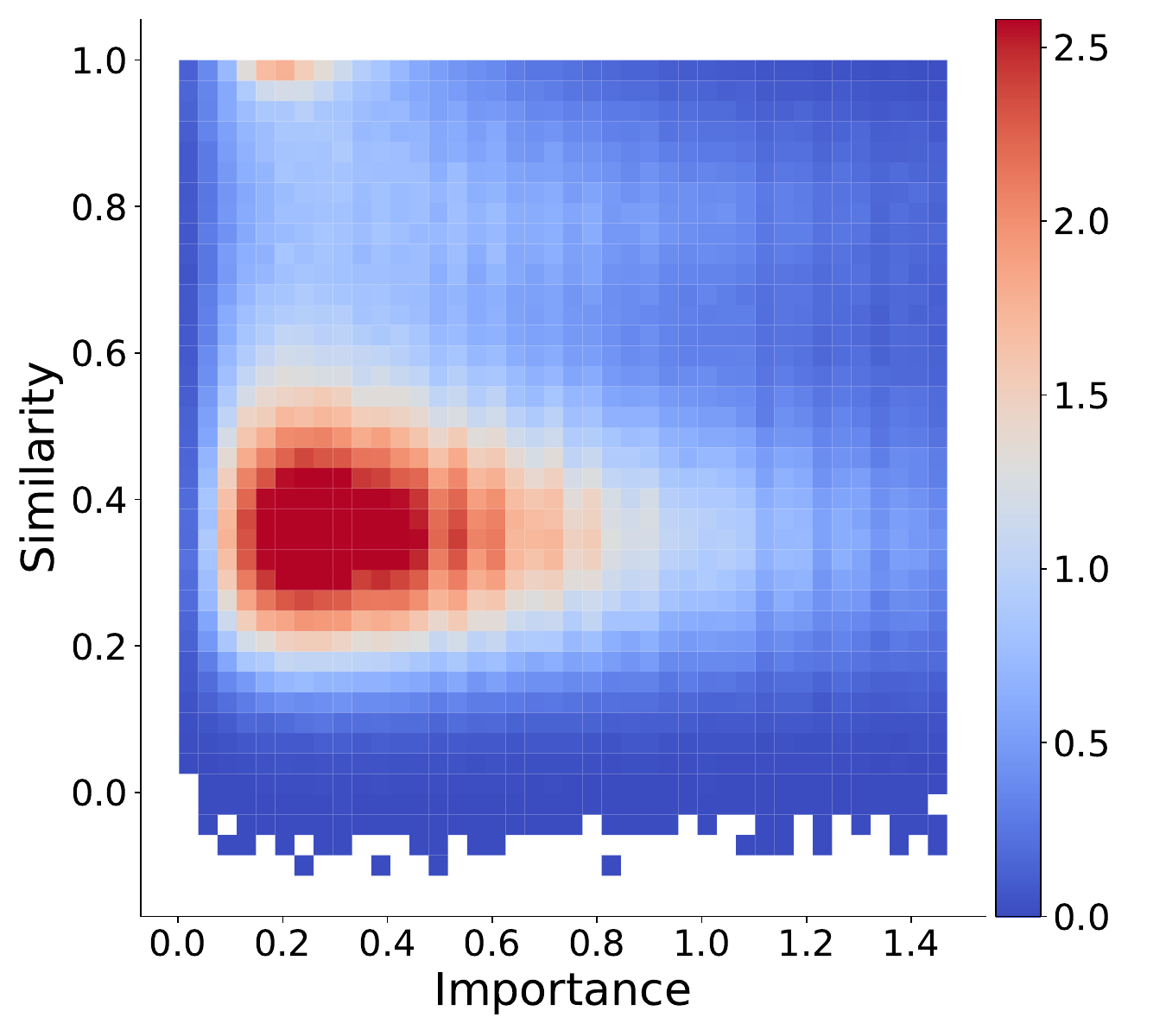}
        \caption{Layer 16}
    \end{subfigure}
    
    \begin{subfigure}{0.3\textwidth}
        \centering
        \includegraphics[width=\linewidth]{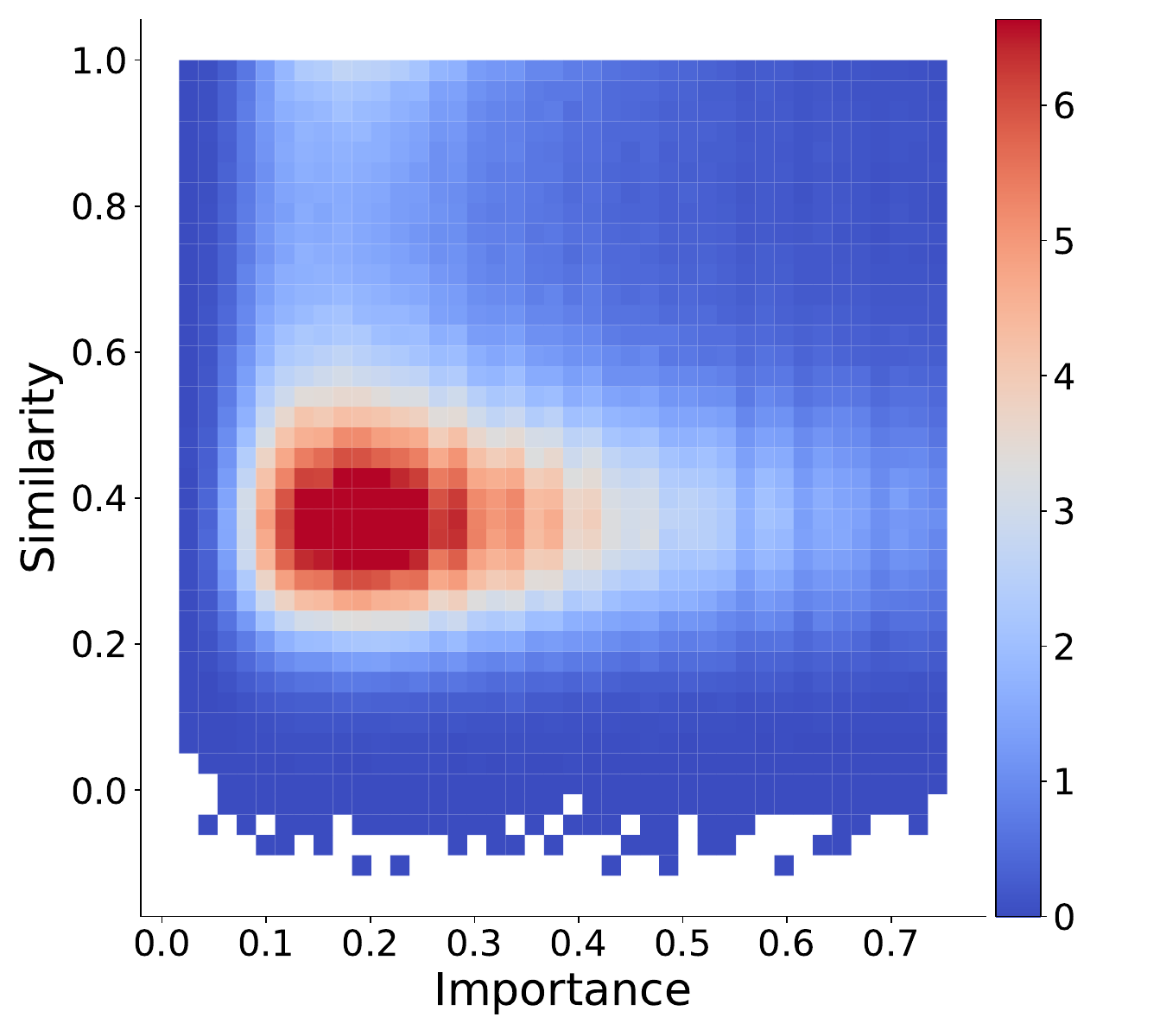}
        \caption{Layer 25}
    \end{subfigure}
    \begin{subfigure}{0.3\textwidth}
        \centering
        \includegraphics[width=\linewidth]{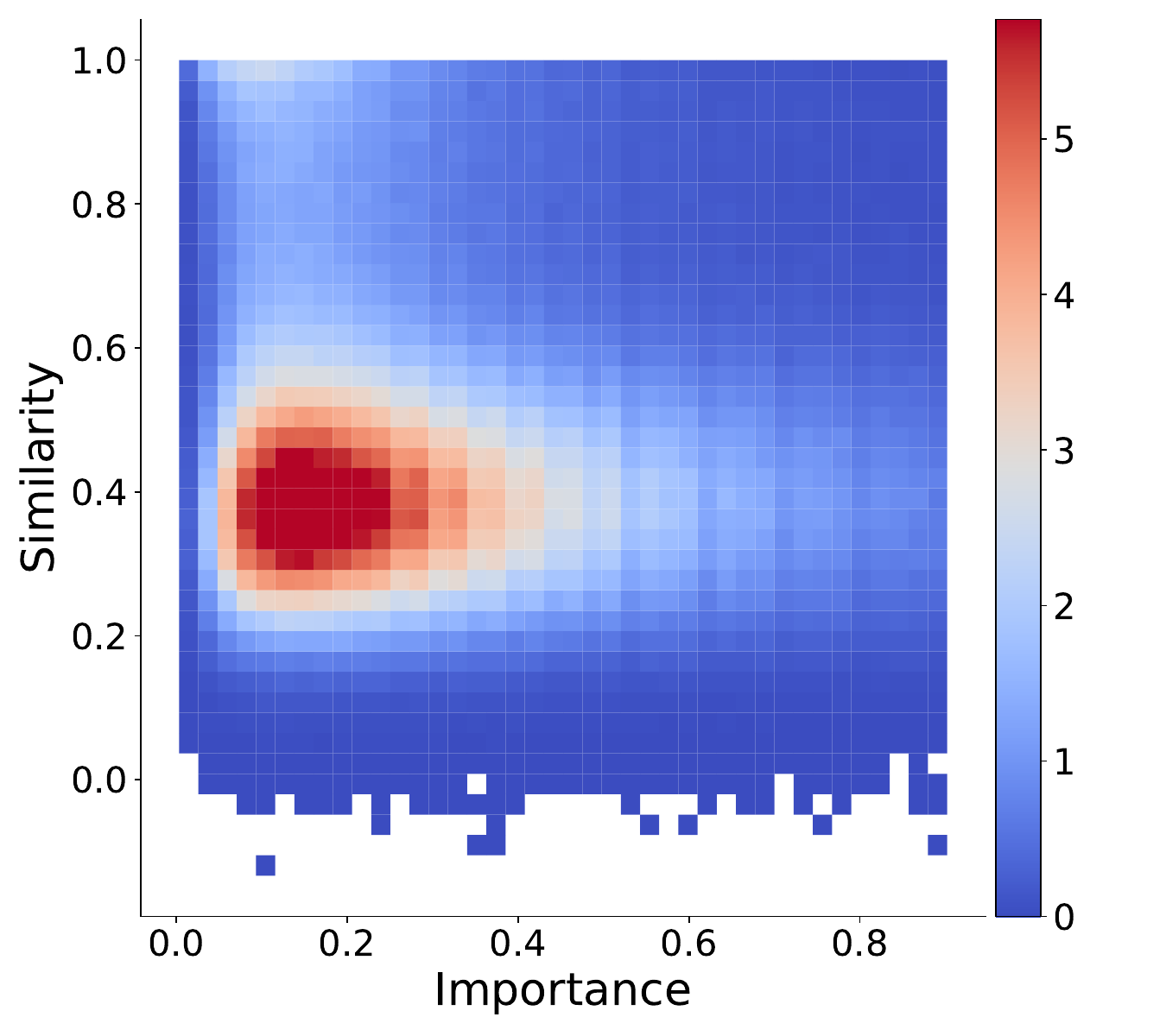}
        \caption{Layer 26}
    \end{subfigure}
    \begin{subfigure}{0.3\textwidth}
        \centering
        \includegraphics[width=\linewidth]{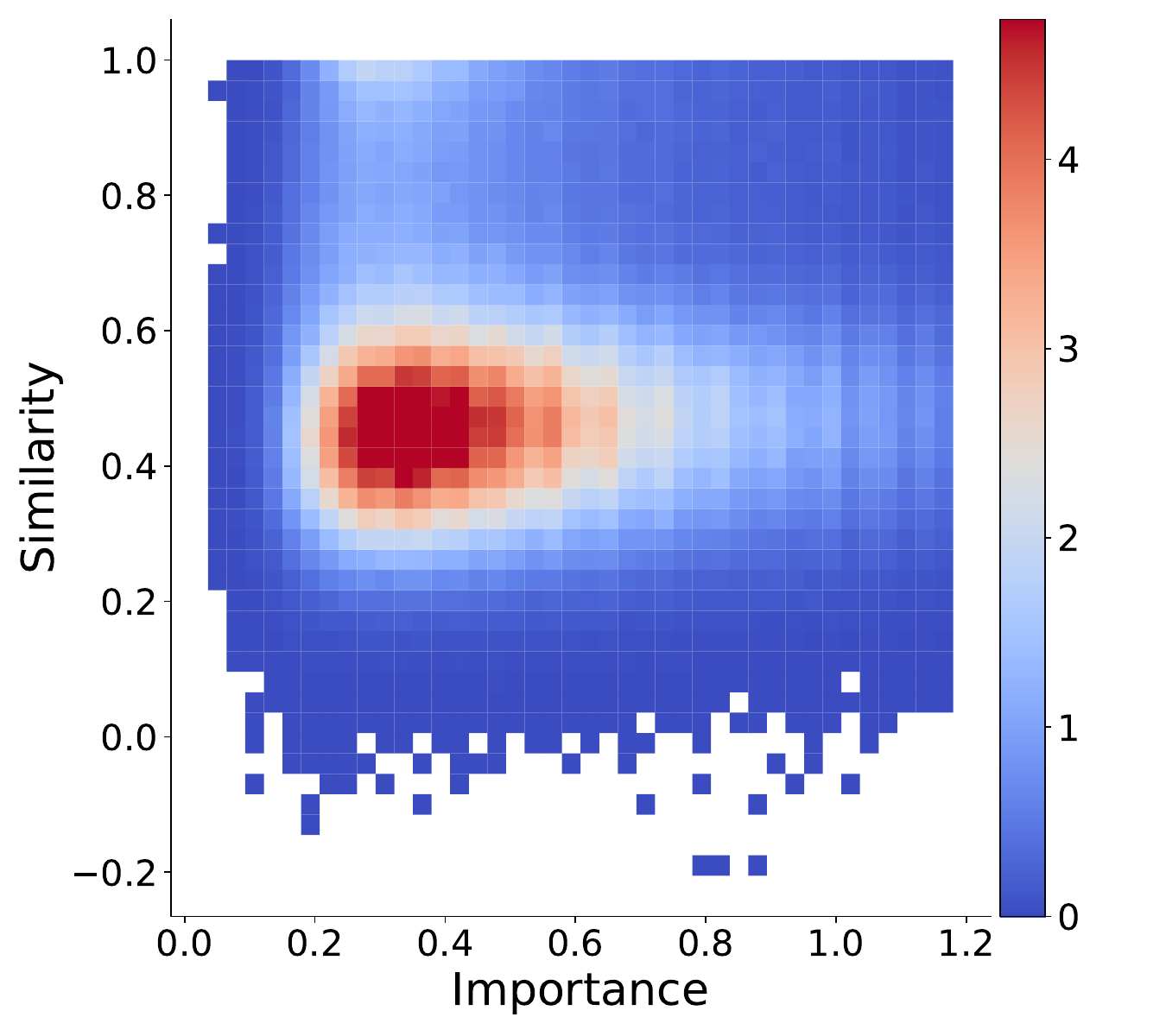}
        \caption{Layer 27}
    \end{subfigure}
    
    \caption{Importance-similarity joint-distribution of different layers, with color intensity denoting distribution frequency.}
    \label{fig:sim_imp_joint}
\end{figure*}

We visualize the joint distribution of token importance and similarity across different layers of Llava-Video-7B. As shown in Figure~\ref{fig:sim_imp_joint}, it can be observed that in the shallow layers of the model, a significant number of tokens exhibit both high similarity and high importance values. \name can effectively compress these tokens. This phenomenon becomes less apparent in the deeper layers of the model, supporting our design choice of performing token merging in the shallow layers of the model.

\section{Additional Discussion on Related Works}
\label{sec:appendix/related_work}

Prior works have also explored token merging in image-based tasks~\citep{Jian2023EVLGen,zhong2024aim,shang2024prumerge}.
For instance, while \name adopts an $O(N)$ \textit{temporal} merging strategy, EVLGen~\citep{Jian2023EVLGen} performs $O(N^2)$ \textit{spatial} merging via bipartite matching among tokens. 
AIM~\citep{zhong2024aim} similarly adopts bipartite-matching-based merging prior to the first layer of the LLM, followed by a token pruning process in subsequent LLM layers, ultimately reducing the number of visual tokens to zero.
LLaVA-Prumerge~\citep{shang2024prumerge} first prunes tokens at the output of the visual encoder and then merges the pruned tokens into the top-$k$ most similar remaining tokens.
In all these methods, the similarity computation incurs a complexity of $O(N^2)$.
Although the computational efficiency of is comparable at the image scale ($N \approx 256$), our method scales more effectively to video scenarios where $N$ can reach 10K to 1M tokens.

\section{Limitation and Future Works}

While \name demonstrates significant improvements in token reduction and efficiency for video LVLMs, certain challenges remain for future work.
First, the similarity-based merging process can be further refined to better handle highly diverse or complex video content, minimizing potential information loss.
Second, the reliance on pre-defined similarity and importance metrics calls for the development of adaptive and task-specific strategies to improve generalization across diverse scenarios.
Future work will focus on designing more robust similarity measures and integrating \name with advanced token-efficient architectures. 

\end{document}